\documentclass[manuscript,screen]{acmart}

\usepackage{multirow}
\usepackage[most]{tcolorbox}
\usepackage{wrapfig}

\AtBeginDocument{%
  }

\setcopyright{acmlicensed}
\copyrightyear{2018}
\acmYear{2018}
\acmDOI{XXXXXXX.XXXXXXX}
\acmConference[Conference acronym 'XX]{Make sure to enter the correct
  conference title from your rights confirmation email}{June 03--05,
  2018}{Woodstock, NY}
\acmISBN{978-1-4503-XXXX-X/2018/06}

\begin{document}

%%
%% The "title" command has an optional parameter,
%% allowing the author to define a "short title" to be used in page headers.

%\title{PlaceReasoner: Reasoning-Driven VLSI Macro Placement and an Open End-to-End Benchmark}

\title{PlaceReasoner-Beta: Reasoning-Driven Macro Placement and Benchmarking}

%\title{PlaceReasoner-Pro: improve Reasoning-Driven macro placment through domain specific finetuning}

%%
%% The "author" command and its associated commands are used to define
%% the authors and their affiliations.
%% Of note is the shared affiliation of the first two authors, and the
%% "authornote" and "authornotemark" commands
%% used to denote shared contribution to the research.
\author{Qiufeng Li}
\email{qiufeng.li@gwu.edu}
\affiliation{%
  \institution{The George Washington University}
  \city{DC}
  % \state{Ohio}
  \country{USA}
}

\author{Chengxuan Wang}
\affiliation{%
  \institution{University of California, Los Angeles}
  \city{Los Angeles}
  \country{USA}}
\email{chengxuan99@ucla.edu}

\author{Rongqian Chen}
\affiliation{%
  \institution{The George Washington University}
  \city{DC}
  \country{USA}
}

\author{Quan Cheng}
\affiliation{%
  \institution{Brown University}
  \city{Providence}
  \country{USA}
}

\author{Yihui Ren}
\affiliation{%
  \institution{Brookhaven National Laboratory}
  \city{Upton}
  \country{USA}
}

\author{Chia-Tung Ho}
\affiliation{%
 \institution{NVIDIA Corp}
 \city{Santa Clara}
 \state{CA}
 \country{USA}}

\author{David Z. Pan}
\affiliation{%
  \institution{UT Austin}
  \city{Austin}
  \state{Texas}
  \country{USA}}

\author{Tian Lan}
\affiliation{%
  \institution{The George Washington University}
  \city{DC}
  \country{USA}}

\author{Weidong Cao}
\correspondingauthor
\affiliation{%
  \institution{The George Washington University}
  \city{DC}
  \country{USA}}
\email{weidong.cao@gwu.edu}

%%
%% By default, the full list of authors will be used in the page
%% headers. Often, this list is too long, and will overlap
%% other information printed in the page headers. This command allows
%% the author to define a more concise list
%% of authors' names for this purpose.
\renewcommand{\shortauthors}{Qiufeng Li et al.}

%%
%% The abstract is a short summary of the work to be presented in the
%% article.
\begin{abstract}

Automated macro placement remains a fundamental challenge in VLSI physical design. Despite decades of research, existing approaches predominantly optimize hand-crafted proxy objectives, such as estimated wirelength, and typically produce placements through one-shot numerical optimization, limiting their ability to incorporate visual layout context, codified design expertise, and downstream physical-design feedback in a unified loop.
We present \textbf{PlaceReasoner-Beta}, a verifier-guided multi-agent framework that reformulates macro placement as a closed-loop reasoning problem rather than black-box optimization. 
A vision-language model (VLM) planner generates candidate placements from the floorplan image, macro specifications, and connectivity structure; a geometric verifier enforces physical legality and expert placement principles; a physical verifier refines candidates using early implementation feedback; and a post-route optimizer further improves promising layouts using final PPA.
To enable reproducible evaluation, we introduce \textbf{PlaceReasoner-Bench}, a fully open end-to-end benchmark built from open RTL designs, EDA tools, and technology libraries. 
It comprises 8 designs at two aspect ratios, yielding 16 tasks with fixed floorplans and I/O assignments, so methods differ only in macro positions and orientations and are evaluated using routed PPA and DRC rather than pre-route proxies.
Across the benchmark, PlaceReasoner-Beta achieves the best timing among DRC-clean methods on all square tasks, reducing post-route TNS by 61.2\% at $1{:}1$ and 53.0\% at $2{:}1$ relative to the classical baseline field. 
It also shortens routed wirelength on most designs despite never explicitly optimizing it, demonstrating that reasoning over spatial structure under physical-design feedback can improve end-to-end layout quality beyond proxy-objective optimization.
Code and benchmark are available at \url{https://github.com/fenggev56/PlaceReasoner-Beta}.
\end{abstract}

%%
%% The code below is generated by the tool at http://dl.acm.org/ccs.cfm.
%% Please copy and paste the code instead of the example below.
%%


%%
%% Keywords. The author(s) should pick words that accurately describe
%% the work being presented. Separate the keywords with commas.

\received{20 February 2007}
\received[revised]{12 March 2009}
\received[accepted]{5 June 2009}

%%
%% This command processes the author and affiliation and title
%% information and builds the first part of the formatted document.
\maketitle

\section{Introduction}

Macro placement is a critical stage in modern very-large-scale integrated circuit (VLSI) physical design, where hundreds to thousands of macros (i.e., large circuit blocks) must be arranged within a constrained floorplan to optimize power, performance, and area (PPA) while satisfying physical requirements such as die dimensions, aspect ratio, and I/O locations~\cite{physical_design}. 
Unlike standard-cell placement, macro placement is dominated by large, heterogeneous objects and long-range geometric interactions, making placement decisions highly consequential to downstream routing, timing, and design-rule closure. 
In practice, experienced designers therefore iteratively inspect layouts, reason about connectivity and physical constraints, and refine macro positions and orientations based on feedback from downstream physical-design tools~\cite{hier_rtl}.
However, the increasing scale and complexity of modern designs, coupled with aggressive time-to-market requirements, make such expert-driven iteration increasingly difficult to sustain, motivating extensive research on automated macro placement methods~\cite{challenge}.
\textcolor{black}{Existing approaches typically formulate macro placement as a constrained optimization problem and employ classical optimization, analytical methods, gradient-based techniques, or reinforcement learning to search the placement space~\cite{dreamplace,nature, macro_v1, macro_v2, macro_v3, macro_v4, autodmp, rtl_mp, chipformer, sa,macro_v5}}. 
While these methods have substantially advanced automated physical design, they remain limited in their ability to reason about the full design context and to optimize for end-to-end physical-design quality.
Three challenges are particularly important.

\textbf{First, existing methods predominantly optimize proxy objectives rather than end-to-end design quality.} Most macro placement approaches rely heavily on metrics such as half-perimeter wirelength (HPWL) as optimization objectives~\cite{rtl_mp}. 
Although computationally efficient, HPWL is only an indirect surrogate for final chip quality and does not fully capture routing congestion, timing, macro orientation, pin accessibility, or design-rule violations.
Consequently, minimizing a proxy objective does not necessarily produce a placement that remains effective after detailed placement and routing. 
Closing this gap requires placement frameworks that can reason from intermediate layout characteristics to downstream physical-design outcomes.
\textbf{Second, existing methods have limited ability to perform multimodal reasoning and leverage domain knowledge.} Expert placement is not simply numerical optimization: designers simultaneously interpret design specifications, connectivity information, and visual layout structures; identify problematic patterns such as congested channels or poorly oriented macros; and apply placement heuristics accumulated through design experience. 
Conventional optimization algorithms, in contrast, largely treat the placement engine as a black-box search procedure. 
They provide limited mechanisms for integrating heterogeneous design information, explicitly reasoning about visual structures, or explaining why a particular placement should be modified.
\textbf{Third, existing methods face substantial challenges in efficient and flexible design-space exploration.} Evaluating a candidate placement with realistic physical-design flows can require computationally expensive placement, routing, timing, and design rule constraint (DRC) analysis. 
Yet optimization-based approaches may require hundreds or thousands of search iterations, many of which provide limited actionable information for subsequent decisions. 
Furthermore, many methods are designed around a fixed floorplan geometry and must be substantially re-optimized when the die dimensions or aspect ratio changes. Such rigidity is problematic because practical floorplanning frequently evolves in response to PPA, packaging, I/O, and manufacturing constraints.

These challenges motivate a shift from \textit{placement as black-box optimization} toward \textit{placement as iterative, tool-grounded reasoning}. Recent advances in large language and vision-language models (LLM/VLM)~\cite{gpt_web, large_survey, cogagent, ui-tars, qwen2-vl, spatialvlm} provide an opportunity to develop agentic systems that can interpret multimodal design information, decompose complex placement tasks, invoke specialized tools, inspect their outputs, and iteratively revise decisions.
However, realizing this opportunity for VLSI macro placement requires more than simply applying a foundation model to generate coordinates. 
An effective framework must combine domain-specific placement knowledge, visual reasoning, geometric verification, EDA-tool interaction, and downstream physical-design feedback within a reproducible optimization loop.

This work presents \textbf{PlaceReasoner-Beta}, a multi-agent vision-language framework for reasoning-driven VLSI macro placement. Given a floorplan image, macro descriptions, die/core specifications, and design connectivity, PlaceReasoner-Beta coordinates specialized agents that collaboratively generate, verify, and refine candidate placements. 
A planner agent proposes placement strategies and candidate layouts; a geometric checker evaluates physical constraints and placement principles; a physical design checker analyzes intermediate implementation feedback and identifies issues that are difficult to capture with simple placement objectives; and a post-route optimizer uses final PPA and DRC results to guide further refinement. 
All agents share a domain-specific skill pack that encapsulates EDA tool interaction and distilled placement knowledge, enabling the system to reason about layout structures and physical design outcomes rather than directly generating fragile tool commands. 
Through this closed-loop architecture, PlaceReasoner-Beta transforms macro placement from a monolithic optimization problem into an \textbf{inspectable, verifiable, and iterative reasoning process}.

To enable systematic evaluation of agentic macro placement systems, we further introduce \textbf{PlaceReasoner-Bench}, an open end-to-end benchmark designed to evaluate placement quality under realistic downstream physical-design flows. 
Unlike existing benchmarks that predominantly focus on square ($1{:}1$) floorplans, PlaceReasoner-Bench evaluates eight open-source designs under both $1{:}1$ and $2{:}1$ aspect ratios, yielding 16 tasks that explicitly test robustness to changing floorplan geometry in practical design scenarios. 
For each task, the floorplan and I/O pin assignment are fixed, isolating macro positions and orientations as the primary variables. 
More importantly, placement quality is evaluated after a complete place-and-route flow using routed PPA and DRC violations, rather than relying solely on pre-route proxy metrics such as HPWL. 
The benchmark is implemented entirely using open-source EDA tools and technology libraries, including Yosys~\cite{yosys} and OpenROAD~\cite{openroad} with Nangate45~\cite{Nangate45}, enabling reproducible evaluation without commercial tool licenses.

Extensive experiments across the 16 benchmark tasks demonstrate the effectiveness of reasoning-driven placement. 
PlaceReasoner-Beta achieves the best timing among all DRC-clean methods on every $1{:}1$ task, reducing post-route total negative slack by $61.2\%$ at $1{:}1$ and $53.0\%$ at $2{:}1$ relative to classical baselines. 
It is also the only evaluated method that successfully routes all eight elongated ($2{:}1$) designs with no more than two DRC violations. 
Notably, PlaceReasoner-Beta improves routed wirelength on most designs even though wirelength is never directly optimized, suggesting that explicit reasoning over layout geometry combined with downstream physical-design feedback can produce higher-quality end-to-end layouts than optimizing a single proxy objective.

Overall, this work makes two contributions: \textbf{(1) an agentic, multimodal framework that integrates placement reasoning, domain knowledge, geometric verification, and physical-design feedback for VLSI macro placement; and (2) an open, end-to-end benchmark that evaluates agentic placement systems using realistic post-route PPA and DRC outcomes across diverse floorplan geometries.} Together, they establish a foundation for developing and systematically benchmarking agentic AI systems for next-generation physical-design automation.

\section{Background and Related Work}

\subsection{VLSI Physical Design Workflow}

\begin{wrapfigure}{r}{0.60\linewidth}
\centering
 \vskip -12pt
\includegraphics[width=1.0\linewidth]{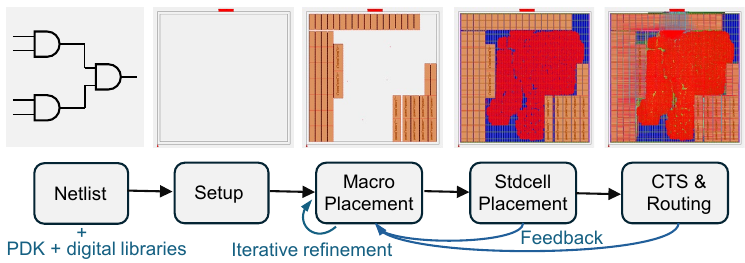}
\vskip -6pt
\caption{
%Macro placement optimization workflow in 
A typical VLSI physical design workflow. Each stage shows its layout
snapshot. Macro placement, the stage we target, fixes the large blocks (orange)
and thereby the whitespace and channels left to stdcell placement (red) and CTS
\& routing (green). The inner blue arc refines a placement from geometric
inspection alone and is cheap, whereas the outer one returns only after a full
downstream run.}
\label{fig:workflow}
\vskip -15pt
\end{wrapfigure}

VLSI physical design translates a synthesized logic netlist into a manufacturable physical layout through a sequence of tightly coupled optimization stages, including floorplanning and macro placement, standard-cell placement, clock-tree synthesis (CTS), and routing~\cite{rtl_mp,vlsi_phy_design} (Fig.~\ref{fig:workflow}). 
Among these stages, \textbf{macro placement is particularly consequential}: the positions and orientations of large circuit blocks, such as SRAMs, often numbering in the hundreds, establish the geometric structure within which all downstream tools must operate and can therefore strongly influence final power, performance, and area (PPA).
In practice, macro placement is not a single-shot optimization problem but an \textbf{iterative reasoning process that combines multimodal design context, placement expertise, and physical-design feedback}. 
Designers first interpret the floorplan context from multiple sources: the die/core geometry and I/O locations from the visual layout, together with the die/core specifications, macro inventory, connectivity, and dataflow information from design descriptions. 
They then apply established placement principles, such as peripheral placement and cluster locality, to reason about pin accessibility, macro connectivity, relative positioning, and available whitespace, generating multiple candidate layouts~\cite{macro_guide}. 
Before invoking expensive EDA tools, designers visually inspect these candidates and repair obvious geometric problems, such as narrow routing channels between macros, excessive whitespace fragmentation, or macro orientations that place pins away from the logic they serve. 
The surviving candidates are progressively evaluated through downstream physical-design stages, where early placement and routing feedback reveals congestion, timing, and other physical constraints.
Clearly inferior candidates are discarded, while promising layouts are iteratively refined until the design approaches physical closure.

%\begin{figure*}[!t]

%\end{figure*}

This workflow highlights a fundamental characteristic of expert macro placement: \textbf{designers optimize not a single numerical objective, but a sequence of interdependent decisions informed by visual structure, domain knowledge, and increasingly expensive physical feedback}.
Although this human-driven process provides no guarantee of global optimality, it remains highly effective because designers can recognize structural problems, adapt placement strategies to different floorplan geometries, and use downstream tool feedback to revise earlier decisions.
Emulating this reasoning loop, rather than merely automating coordinate optimization, is therefore a key opportunity for agentic approaches to VLSI macro placement.

\subsection{Automated Placement Methods and Benchmarks}

Automated macro placement has been widely studied over the past few decades and has evolved along two major directions, \textbf{classical optimization} and \textbf{learning-based methods}, with support from a relatively small set of public benchmarks. 
Despite substantial progress, existing approaches largely formulate macro placement as a numerical search problem and evaluate candidates using proxy objectives. 
This paradigm differs fundamentally from expert placement, where designers jointly reason over layout geometry, connectivity, design constraints, and downstream physical-design feedback. 
We review these approaches below and highlight three gaps that motivate \textbf{PlaceReasoner-Beta}: \textit{(1) proxy-driven rather than end-to-end optimization, (2) limited multimodal and iterative reasoning, and (3) insufficiently realistic and reproducible benchmarking}.

\noindent\textbf{Classical methods.}
Classical macro-placement approaches~\cite{dreamplace,macro_v1,macro_v2,macro_v3,macro_v4,maskplace,rtl_mp,macro_v5} can be broadly categorized into three families. \textit{Packing-based methods} represent macro-to-macro geometric relationships using compact structures such as sequence pairs and B*-trees, enabling efficient exploration of feasible floorplans~\cite{macro_v5}. 
\textit{Analytical methods}~\cite{autodmp,eplace-ms,ntuplace4h,effective_todaes} formulate placement as continuous optimization, minimizing differentiable objectives that typically combine smoothed wirelength with density and overlap penalties; DREAMPlace~\cite{dreamplace} is a representative example.
\textit{Hierarchy-aware methods}~\cite{rtl_mp,hierarchy_v1,hiercrchy_v2,hierarchy_v3,hierarchy_v4} exploit design hierarchy and architectural structure by grouping logically related blocks and incorporating these relationships into mixed-size placement to improve routability.

Despite their different formulations, these methods share a common optimization paradigm: \textbf{they rely primarily on hand-crafted numerical objectives, with wirelength serving as the dominant proxy for placement quality}. 
Such objectives are attractive because they are computationally efficient, but they incompletely capture the factors that determine final design quality, including routing congestion, timing, pin accessibility, macro orientation, and DRC violations. 
Consequently, a placement that is optimal under the surrogate objective may perform poorly after detailed placement and routing.
Moreover, these methods generally produce placements through monolithic numerical optimization, providing limited mechanisms for interpreting a layout, identifying structural defects, or incorporating qualitative observations and downstream physical-design feedback into subsequent decisions.
\textbf{The first gap, therefore, is the disconnect between proxy optimization and end-to-end physical-design quality.}

\noindent\textbf{Learning-based methods.}
Learning-based approaches seek to reduce dependence on manually designed optimization moves and cost functions by learning placement strategies from data or interaction~\cite{learn_local}. 
Google introduced deep reinforcement learning (RL) to sequentially place macros based on learned policies~\cite{nature}. MaskPlace~\cite{maskplace} formulates placement as a visual representation-learning problem and improves wirelength while enforcing non-overlap constraints, while ChiPFormer~\cite{chipformer} employs offline RL and adapts to unseen designs to improve placement efficiency. These methods demonstrate that learned representations and policies can capture placement patterns beyond explicit optimization heuristics.
However, their reward functions remain predominantly tied to wirelength or related placement-level objectives, and the learned policies do not explicitly reason over the heterogeneous information available to human designers, such as floorplan images, textual design specifications, connectivity, and physical-design feedback.
%\textbf{Reasoning-based methods.} 
Recent advances in vision-language models (VLMs)~\cite{CLIP,qwen2-vl} provide a promising opportunity to bridge this gap. 
Modern VLMs have demonstrated increasingly strong multimodal and spatial reasoning capabilities~\cite{internvl3,spatialvlm,layout-3d}, including object localization, relative-position reasoning, alignment, and geometric relationship understanding.
These capabilities are particularly relevant to macro placement, where decisions depend on both visual layout structure and semantic relationships among circuit blocks. 
VeoPlace~\cite{veoplace} represents an early step in this direction by using a VLM to shape the search space of an underlying placer, such as an analytical engine, rather than directly generating and refining macro layouts. 
MAGE~\cite{mage} further explores a multi-agent architecture based on VLMs for physical design. 
However, its dependence on commercial EDA tools and a proprietary technology node limits independent reproduction and systematic evaluation.
More fundamentally, existing VLM-based approaches have not yet fully established a \textbf{closed-loop reasoning paradigm for macro placement}. 
An effective agentic placer should do more than predict coordinates: it should interpret the design context, propose candidate layouts, inspect their geometric validity, obtain physical-design feedback, diagnose problems, and revise the placement accordingly. 
\textbf{The second gap is therefore the lack of an agentic, multimodal, and feedback-driven framework that explicitly models the iterative reasoning process of expert macro placement.} 
PlaceReasoner-Beta addresses this gap by placing VLM-based agents directly in the decision loop, where specialized agents collaboratively propose, inspect, verify, and refine layouts using both geometric reasoning and downstream physical-design feedback.

\noindent\textbf{Benchmarks.}
The evaluation infrastructure for macro placement is considerably less mature than the algorithmic landscape. 
ChiPBench~\cite{chipbench} provides an open-source end-to-end physical-design flow across 20 circuits, but approximately half contain no macros, while its floorplans are relatively loosely constrained, using around $30\%$ utilization compared with the $60{\sim}70\%$ commonly encountered in practice.
RTL-MP~\cite{rtl_mp,hier_rtl} provides six macro-containing designs with downstream place-and-route (P\&R) scripts, enabling evaluation beyond placement-only metrics. However, each design is evaluated at only a single aspect ratio, and its flow depends on commercial EDA tools and a proprietary 12\,nm foundry technology, limiting independent reproduction. 
PDAgent-Bench~\cite{pdagent} evaluates agents across ten full physical-design tasks, but targets general flow-level agent competence rather than isolating the quality of macro placement itself.
These limitations expose a \textbf{third gap: the lack of an open, realistic, and geometrically diverse benchmark specifically designed to evaluate macro-placement reasoning}. In particular, existing benchmarks provide limited variation in floorplan geometry, rely in some cases on commercial or proprietary implementation flows, and often emphasize placement-level proxies or broad flow-level success rather than the quality of the resulting routed design. 
A meaningful benchmark for agentic macro placement should therefore (i) contain realistic macro-intensive designs, (ii) test robustness across different floorplan geometries, (iii) provide a fully reproducible open-source implementation flow, and (iv) evaluate placements using downstream PPA and DRC outcomes rather than relying solely on pre-route proxies.

\textbf{PlaceReasoner-Beta addresses these three gaps jointly.} It introduces a reasoning-driven multi-agent framework in which VLM agents use multimodal design context and domain knowledge to generate, inspect, and iteratively refine macro placements under physical-design feedback. In parallel, \textbf{PlaceReasoner-Bench} provides an open end-to-end benchmark with multiple floorplan aspect ratios and post-route evaluation based on PPA and DRC outcomes. 
Together, the framework and benchmark shift macro placement from \textit{single-objective coordinate optimization} toward \textit{iterative, multimodal, and physically grounded reasoning}, while providing a reproducible foundation for systematically evaluating future agentic approaches to VLSI physical design.

\section{PlaceReasoner-Beta}

% \subsection{}

\subsection{Framework Overview}

\begin{figure*}[!t]
\centering
\includegraphics[width=1\linewidth]{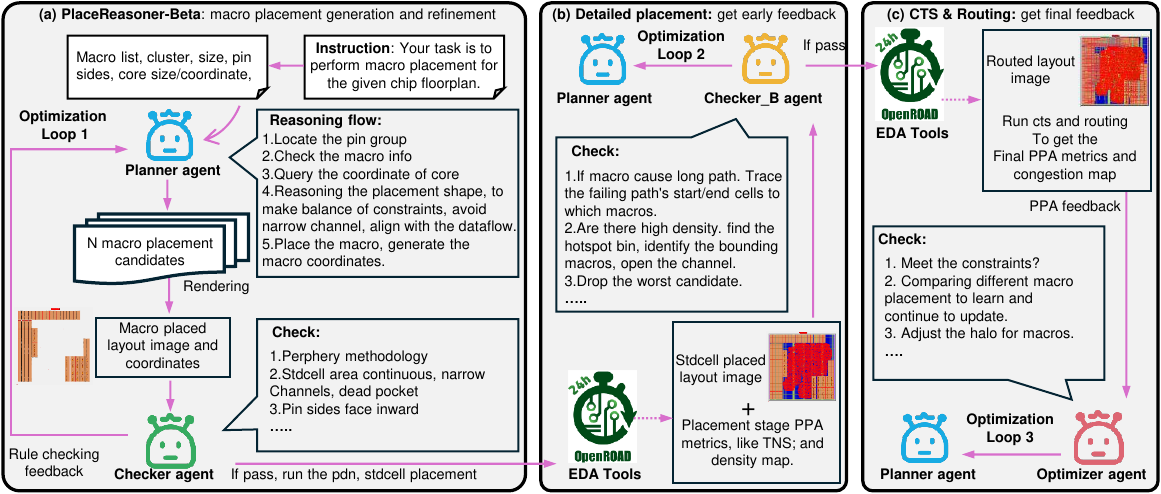}
\vskip -6pt
\caption{PlaceReasoner-Beta framework overview. It has four agents: Planner, Geometric Checker, Physical Checker, and Post-route Optimizer. These agents formulate three stages essential in physical design: (a) PlaceReasoner-Beta for macro placement generation and refinement, (b) detailed placement by getting early feedback, and (c) CTS \& routing by getting final feedback.  }
\vskip -6pt
\label{fig:placereasoner}
\end{figure*}

PlaceReasoner-Beta is an agentic framework for VLSI macro placement that treats placement as a \textbf{closed-loop reasoning and verification problem} rather than one-shot coordinate optimization. 
Given a floorplan image, macro specifications, I/O locations, connectivity information, and natural-language design objectives, PlaceReasoner-Beta reasons about the position and orientation of each macro by combining visual-spatial reasoning with codified placement knowledge. 
The framework is designed around a key observation from expert physical design: high-quality placement emerges through successive rounds of proposal, inspection, physical evaluation, and refinement, rather than from optimizing a single placement objective.
This design introduces three challenges that an agentic placement framework must address.
\textbf{First, physical feedback is expensive and delayed.} Meaningful placement quality is often revealed only after downstream stages such as standard-cell placement, CTS, and routing, making exhaustive evaluation of candidate layouts prohibitively expensive. 
\textbf{Second, EDA interaction creates a reasoning-to-execution gap.} Placement intent expressed at a high level must ultimately be translated into precise EDA commands, while tool feedback arrives as heterogeneous reports, metrics, and physical maps that must be interpreted before they can inform the next placement decision. \textbf{Third, macro placement is inherently multi-objective.} Timing, congestion, wirelength, PPA, routability, and geometric feasibility are tightly coupled and often conflicting, requiring the system to reason about tradeoffs rather than optimize a single scalar reward.

PlaceReasoner-Beta addresses these challenges through a \textbf{multi-agent, coarse-to-fine optimization framework} that mirrors the iterative workflow of physical designers (Fig.~\ref{fig:placereasoner}). Four specialized agents collaborate across three progressively more expensive feedback loops. 
The \textit{Planner} generates diverse placement candidates from multimodal design context; \textit{Geometric Checker} performs fast geometric verification and identifies violations of hard constraints and placement principles; \textit{Physical Checker} interprets early physical-design feedback after standard-cell placement and recommends targeted refinements; and the \textit{Post-route Optimizer} uses final routed PPA and physical metrics to refine the most promising candidates. 
This staged architecture deliberately postpones expensive post-route evaluation until poor candidates have been eliminated, improving the efficiency of design-space exploration.

All agents share a reusable \textit{skill pack} that abstracts EDA-tool interaction and encodes placement expertise, including templated P\&R scripts, layout and density extraction, verification procedures, and distilled expert heuristics. 
This abstraction separates \textbf{reasoning from tool execution}: agents reason about the desired layout and its physical consequences, while the skill pack reliably translates those decisions into executable EDA operations and returns structured feedback. 
Because the benchmark designs do not provide explicit dataflow annotations, PlaceReasoner-Beta additionally derives connectivity-based macro clusters and exposes them to the Planner as soft placement constraints, encouraging locality among logically related macros~\cite{rtl_mp}.
Importantly, PlaceReasoner-Beta is \textbf{tuning-free with respect to the underlying foundation model}. 
The placement principles, verification rules, feedback interpretation procedures, and skill pack are framework artifacts rather than learned parameters. They can therefore be inspected, corrected, extended, or replaced without retraining the backbone model.
This property makes the framework particularly suitable for systematic benchmarking: improvements can be attributed to changes in the reasoning framework rather than to hidden model-specific training.
To enable such evaluation, we develop \textbf{PlaceReasoner-Bench}, an open end-to-end benchmark specifically designed to measure macro-placement quality under realistic downstream physical-design conditions. The framework and benchmark are developed together: PlaceReasoner-Beta provides the agentic methodology for generating and refining layouts, while PlaceReasoner-Bench provides a controlled environment in which different placement strategies can be compared using identical downstream flows and post-route outcomes.

\subsection{Agent Design and Closed-Loop Optimization}

PlaceReasoner-Beta organizes placement refinement into three feedback loops, progressing from inexpensive geometric verification to increasingly realistic physical-design evaluation.
This coarse-to-fine strategy allows the system to spend computational resources in proportion to candidate quality.

\noindent\textbf{Planner: multimodal placement generation.}
The Planner is responsible for generating diverse candidate layouts from the available design context. 
Its input includes the macro list and dimensions, core boundaries, I/O locations, connectivity-derived clusters, and natural-language placement objectives. 
The Planner prompt encodes established placement principles, spatial reasoning strategies, and a structured output format for executable macro coordinates and orientations (Fig.~\ref{fig:placereasoner}(a)).
Each macro is represented using structured geometric attributes, position, orientation, and pin-facing direction, together with a rendered floorplan image. 
The structured representation provides precise information required for EDA execution, while the image exposes global spatial relationships that are difficult to capture through coordinates alone. 
The Planner jointly reasons about macro geometry, cluster locality, pin accessibility, peripheral placement, whitespace, and orientation, and produces $N$ candidate placements. 
These candidates are then translated into native EDA commands through the shared skill pack.

\noindent\textbf{Geometric Checker: geometric verification and repair.}
Geometric Checker provides a fast verification layer before any expensive physical-design evaluation. 
It checks each candidate against hard geometric constraints, including core-boundary violations, macro overlap, minimum spacing, peripheral-placement requirements, and pin-side compatibility.
Rather than returning only a binary valid/invalid decision, Geometric Checker generates diagnostic feedback describing the structural cause of each violation. 
The Planner uses this feedback to revise the candidate in Loop~1 (Fig.~\ref{fig:placereasoner}(a)).
Because VLM-generated coordinates may contain small residual overlaps even after iterative correction, PlaceReasoner-Beta includes a deterministic legalization procedure as a final safeguard. Legalization resolves remaining overlaps while minimizing perturbation from the intended placement, ensuring that numerical imprecision in coordinate generation does not unnecessarily consume expensive physical-design evaluations.

\noindent\textbf{Physical Checker: early physical verification.}
Candidates that pass geometric verification are propagated through standard-cell placement to obtain inexpensive but informative physical-design feedback. 
Physical Checker analyzes early indicators such as estimated PPA, cell-density distributions, congestion-related features, and timing information.
Instead of treating these measurements solely as scalar rewards, the checker interprets them spatially and diagnostically; for example, identifying congested regions, problematic macro channels, or placement structures likely to create timing bottlenecks. 
It then provides targeted recommendations to the Planner for Loop~2 (Fig.~\ref{fig:placereasoner}(b)).
This stage serves as the bridge between geometric reasoning and physical optimization: candidates that are geometrically legal but physically unfavorable can be rejected or revised before incurring the substantially higher cost of CTS and routing.

\noindent\textbf{Post-route Optimizer: end-to-end refinement.}
The final loop evaluates only the most promising candidates through CTS and routing and uses the resulting physical-design outcomes to guide further refinement. 
The Post-route Optimizer considers routed PPA, timing, congestion, and DRC-related feedback and identifies the placement structures responsible for remaining deficiencies. 
It then requests targeted placement modifications from the Planner rather than restarting the search from scratch (Fig.~\ref{fig:placereasoner}(c)). 
Iteration continues until the improvement converges or the refinement budget is exhausted.
The three loops therefore establish a progressively more expensive decision
process, screening candidates first on geometric feasibility, then on early
physical quality, and only finally on post-route PPA and DRC.
This architecture enables PlaceReasoner-Beta to selectively use expensive physical-design feedback while preserving the iterative inspection and revision characteristics of expert placement.

\noindent\textbf{Placement representation and spatial reasoning.}
A placement is represented at two complementary levels. 
At the structured level, each macro is described by its location, dimensions, orientation, and pin-facing direction, enabling deterministic translation into EDA commands. 
At the visual level, the complete floorplan is rendered as an image, allowing a VLM to reason about global spatial relationships, relative distances, whitespace, clusters, and boundary interactions. 
The combination avoids relying exclusively on either symbolic coordinates or visual perception: structured attributes provide execution precision, while the rendered layout provides the spatial context needed for reasoning.

\noindent\textbf{Connectivity-based macro clustering.}
The open RTL designs used in our benchmark do not provide complete dataflow annotations. 
To provide the Planner with useful structural context without introducing manually curated placement labels, we derive soft macro clusters directly from netlist connectivity. 
Related macros are grouped according to their connectivity structure, and the resulting clusters are supplied to the Planner as placement preferences rather than hard constraints. 
This encourages dataflow locality while allowing the agent to override clustering when geometric or physical considerations suggest a different arrangement.

\noindent\textbf{Macro overlap legalization.}
VLM-generated coordinates are often close to legal but may contain small numerical errors; a single overlapping macro pair is sufficient to invalidate an otherwise promising candidate. PlaceReasoner-Beta therefore applies force-directed legalization after candidate generation.
Overlapping macro pairs are displaced along the axis of smaller overlap until the required $2\mu$m spacing is restored, while weak restoring forces pull macros toward their intended positions and all macros remain clamped within the core boundary. 
If overlaps remain after the iteration budget, a greedy fallback places macros in descending area order at the nearest legal positions. 
This deterministic safeguard preserves the Planner's spatial intent while removing the burden of exact overlap arithmetic from the VLM.

\subsection{Benchmark Development}

\begin{wraptable}{r}{0.47\linewidth}
\vspace{-\intextsep}
\centering
\footnotesize
\setlength{\tabcolsep}{3pt}
\caption{Composition of PlaceReasoner-Bench. \#Types counts distinct macro
types, Clk is the target clock period, and M/Core is macro area as a fraction of
core area. ``--'' denotes a quantity without aggregation.}
\label{tab:pr_bench}
\vskip -6pt
\begin{tabular}{@{}l ccc cc@{}}
\toprule
Design & \#Macros & \#Clusters & \#Types & Clk (ns) & M/Core (\%) \\
\midrule
ariane133      & 132 & 14 & 1 & 4.0 & 34.4 \\
ariane81       &  81 &  8 & 6 & 4.0 & 36.0 \\
bp\_be         &  10 &  3 & 3 & 2.6 & 36.7 \\
bp\_fe         &  11 &  4 & 3 & 1.8 & 45.6 \\
swerv\_wrapper &  28 &  3 & 3 & 2.0 & 44.1 \\
vga\_lcd       &  62 & 12 & 3 & 2.0 & 40.5 \\
ethernet       &  64 &  2 & 1 & 8.0 & 14.2 \\
VeriGPU        &  12 &  1 & 1 & 1.2 & 30.5 \\
\midrule
Total          & 400 & 47 & -- & -- & -- \\
\bottomrule
\end{tabular}
\vskip -6pt
\end{wraptable}

\textbf{PlaceReasoner-Bench: an open, end-to-end benchmark.}
A central contribution of this work is PlaceReasoner-Bench, which is designed to evaluate \textbf{agentic macro placement rather than generic physical-design automation}. 
The benchmark is constructed entirely from open-source RTL, EDA tools, and technology libraries so that every task and downstream result can be regenerated without commercial licenses (Table~\ref{tab:pr_bench}).
More importantly, the benchmark fixes all factors other than macro placement, enabling controlled attribution of post-route differences to the generated macro coordinates and orientations.
Each benchmark task is generated through a deterministic pipeline. We first synthesize the RTL using Yosys~\cite{yosys} with the open Nangate45 standard-cell library~\cite{Nangate45}.
SRAM macros instantiated by the designs are generated using FakeRAM~2.0, an open memory compiler that provides macro abstracts with physical dimensions and timing characterization. 
This eliminates dependence on proprietary memory compilers, which otherwise prevents independent reproduction of macro-placement experiments.

For each synthesized design, we construct a realistic floorplan targeting approximately $60\%$ core utilization and instantiate two floorplan geometries with aspect ratios of $1{:}1$ and $2{:}1$. I/O pins are assigned once for each task using the automatic pin-placement engine in OpenROAD~\cite{openroad} and then frozen.
Consequently, every method receives the same netlist, core geometry, macro library, and I/O assignment, and the only variables controlled by the placer are macro positions and orientations.
Fixing the pin assignment is particularly important for macro-placement evaluation: because macro orientation should align pin-facing sides with the logic they communicate with, the boundary pin pattern represents a genuine physical constraint rather than an additional optimization degree of freedom.
The benchmark contains eight open-source RTL designs, including \texttt{ariane133}, \texttt{ethernet}, and \texttt{vga\_lcd}. 
Across these designs, the benchmark contains 10--132 macros per design and 400 macros in total, organized into 47 connectivity clusters, with 1--6 macro types per design.
Macro-to-core area ratios range from $14.2\%$ to $45.6\%$, while target clock periods range from $1.2$ to $8.0$~ns. 
Evaluating each design at two aspect ratios yields \textbf{16 end-to-end tasks}, spanning both relatively macro-sparse designs and macro-dominated designs, in which macro arrangement imposes substantially stronger constraints on downstream implementation.

\noindent\textbf{Controlled end-to-end evaluation.}
All stages after macro placement are held fixed across competing methods, including standard-cell placement, CTS, routing, timing repair, technology libraries, and implementation settings. 
Thus, differences in final PPA and DRC outcomes can be attributed to the macro placement itself rather than differences in downstream optimization. 
Each candidate is evaluated through the same complete physical-design flow, making routed PPA and DRC violations the primary measures of placement quality.
The resulting benchmark evaluates properties that are difficult to capture with conventional placement metrics.
In particular, the $1{:}1$ and $2{:}1$ configurations test whether a placement strategy can adapt to substantially different spatial geometries; fixed I/O assignments test whether macros can be oriented and positioned according to pin accessibility; realistic utilization creates meaningful whitespace and routing constraints; and post-route evaluation reveals whether apparently good geometric placements actually translate into high-quality physical implementations. 
\textbf{PlaceReasoner-Bench therefore transforms macro placement from a placement-level coordinate task into a reproducible end-to-end evaluation of physical-design reasoning.}

Together, PlaceReasoner and PlaceReasoner-Bench establish a unified framework for developing and evaluating agentic macro placement: \textbf{the framework provides a closed-loop reasoning mechanism, while the benchmark provides a controlled and reproducible testbed for measuring whether that reasoning produces better physical designs.}

\section{Experimental Setup}

\noindent\textbf{Models.}
We instantiate PlaceReasoner-Beta on three backbones: the open-weight Qwen3-VL-8B~\cite{qwen3} and Qwen3-VL-30B, and the proprietary Claude-Opus-4.8~\cite{opus48}, and evaluate all three on PlaceReasoner-Bench.
Unless otherwise specified, the generation temperature is set to $0.1$.

\noindent\textbf{EDA Flow.}
We use Yosys~\cite{yosys} for logic synthesis, OpenROAD~\cite{openroad} for P\&R, and FakeRAM~2.0~\cite{sa} to generate memory macros, all under the open-source Nangate45 technology~\cite{Nangate45}.
FakeRAM~2.0 is an open-source memory compiler that emits SRAM macro abstracts with physical dimensions and timing characterization, so the generated macros can be placed and routed like foundry memories; this removes any dependence on a commercial PDK or memory compiler and keeps the entire flow reproducible.
Unless stated otherwise, all methods share the same netlist, floorplan, macro library, and P\&R and timing-repair settings.
Thus, they differ only in the generated macro coordinates and orientations.

\noindent\textbf{Benchmark Tasks.}
We evaluate our framework on the proposed PlaceReasoner-Bench (Table~\ref{tab:pr_bench}), which measures end-to-end quality after a complete P\&R flow: its 8 open-source designs, each evaluated at aspect ratios $1{:}1$ and $2{:}1$, form 16 tasks scored by routed PPA, wirelength, and DRC violations rather than by a pre-route wirelength proxy alone.

\noindent\textbf{Baselines.}
On PlaceReasoner-Bench, we compare against four representative existing approaches:
DREAMPlace~4.0~\cite{dreamplace}, a GPU-accelerated analytical placer;
RTL-MP~\cite{rtl_mp}, a hierarchy- and dataflow-aware placer;
ChiPFormer~\cite{chipformer}, an offline RL method; and a
conventional simulated-annealing (SA) packing baseline~\cite{sa}.
Each baseline is run with the hyper-parameters reported in its original
paper~\cite{dreamplace, sa, rtl_mp, chipformer}.
Unless stated otherwise, PlaceReasoner-Beta denotes the full system on Claude-Opus-4.8, while \textsc{Qwen3-VL-8B} and \textsc{Qwen3-VL-30B} denote the same agentic workflow and skill pack running on the corresponding open-weight backbones, so that the gap between them isolates the effect of the backbone.
We also report two controlled ablations, both on Claude-Opus-4.8: \textsc{CWI} removes layout images, and \textsc{CWS} keeps images but disables the skill pack.
For PlaceReasoner-Beta, we set the budget of {Germetric Checker} and {Physical Checker} to 3, the Optimizer budget to 4, and the number of placement candidates to 5.
To account for run-to-run variance, every method is executed 5 times per design, and we report its best run.
%so that all methods receive an identical trial budget.

\noindent\textbf{Evaluation Metrics.}
On PlaceReasoner-Bench, we report routed worst negative slack (WNS), total negative slack (TNS), design-rule-check (DRC) violations, routed wirelength, total power, and chip area.
A run counts only once detailed routing succeeds; timeouts and tool failures are recorded as failed runs rather than replaced with pre-route estimates.
We further report the wall-clock runtime of every method and, for the agentic ones, the total token usage, so that placement quality can be weighed against its compute cost.

\iffalse
\textcolor{red}{WD: which aspects/capabilities to evaluate.}

\noindent\textbf{Evaluation Frameworks.} 
We first evaluate all the methods overall capability on macro placement through evaluating on PlaceReasoner-Bench, the performance represented by the routed PPA.
Better PPA means better macro placement, the method will be stronger. 
We also illustrate the image input is important, by showing the macro orientation. 
For reasoning methods, we provide the variance data, this is the exploration process, we are not just generate one candidate, although vary, reasoning method consistently outperform the classical methods.
For showing the optimizer agent, we show the optimization trajectory, reasoning method can continue to optimize the PPA, but classical methods cannot.

We will add ablation study to justify the image input and human knowledge expertise.

Finally, we report the token usage and model runtime.
\fi

\noindent\textbf{Evaluation Framework.}
We evaluate PlaceReasoner-Beta along four complementary dimensions that capture both \textbf{placement effectiveness} and the mechanisms underlying agentic reasoning. 
\emph{(1) Placement quality}: We evaluate routed PPA and DRC violations on PlaceReasoner-Bench, using post-route outcomes as the primary measure of macro-placement quality rather than pre-route proxy objectives. 
\emph{(2) Visual grounding}: We assess whether explicit layout perception contributes to placement quality through a \textsc{CWI} ablation that removes the floorplan image while preserving the remaining design context, thereby quantifying the benefit of visual reasoning, particularly for macro orientation and spatial relationships. 
\emph{(3) Exploration robustness}: We examine the quality distribution across the five candidates generated in each run, evaluating whether PlaceReasoner-Beta consistently produces strong solutions rather than relying on a single favorable sample. 
\emph{(4) Iterative refinement}: We track the optimization trajectory across successive feedback loops to determine whether the agent can systematically improve placement quality using geometric, early physical, and post-route feedback. 
We additionally report runtime and token consumption to characterize the computational cost of agentic reasoning, and perform a \textsc{CWS} ablation that removes the shared skill pack to isolate the contribution of explicitly encoded placement expertise and EDA-tool knowledge.

\iffalse

We evaluate PlaceReasoner-Beta from four aspects.
\emph{(1) Placement quality}: routed PPA on PlaceReasoner-Bench, where better PPA with fewer DRC violations
means a better macro placement.
\emph{(2) Visual grounding}: whether the layout vision modality is important. Reasoning over
the image lets the model choose better macro orientations, and the \textsc{CWI}
ablation, which removes the image, quantifies how much of the gain it accounts
for.
\emph{(3) Robust exploration}: the spread over the five candidates of each run, where
reasoning methods beat the classical baselines despite that variance.
\emph{(4) Iterative refinement}: the optimizer trajectory, along which reasoning
methods keep improving PPA while the classical methods cannot.
We further report runtime and token usage as the cost of these capabilities, and
ablate the skill pack (\textsc{CWS}) to isolate the contribution of encoded
human expertise.

\fi

% \subsection{Evaluations}

\section{Evaluation Results and Analysis}

\begin{table*}[t]
\centering
\scriptsize
\setlength{\tabcolsep}{3pt}
\caption{Post-route results for four representative designs at the square ($1{:}1$)
aspect ratio; the remaining four designs are reported in the appendix.
``--'' denotes an unavailable metric and failure states indicate that detailed
routing did not complete. Winner \# reports the one-based winner position $k/N$
with $N=5$, and is undefined for the pure-model \textsc{Qwen3-VL-8B} and
\textsc{Qwen3-VL-30B} rows, which report the single selected candidate.
Runtime is macro-placement time for the classical baselines and
summed model-driver time for the agent variants; Token Usage is the total of
model input and output tokens. For ariane133, \textsc{Qwen3-VL-30B} left
18{,}166 residual violations, so its timing numbers are extracted from a
non-converged routing and are not comparable with the clean rows.}
\vskip -6pt
\label{tab:pr_detail_ar1_4}
\begin{tabular}{llcrrrrrrrcc}
\toprule
Design & Method & Winner \# & Route & WNS (ns) & TNS (ns) & DRC & WL ($\mu$m) & Power (W) & Area ($\mu$m$^2$) & Runtime (s)& Token Usage (k) \\
\midrule
ariane133 & RTL-MP & 2/5 & Yes & -0.10 & -4.60 & 0 & 5233982 & 0.265 & 729834 & 127.0 & -- \\
ariane133 & DREAMPlace & 3/5 & Fail: timeout & -- & -- & -- & -- & -- & -- & 11.00 & -- \\
ariane133 & ChiPFormer & 1/5 & Yes & -0.04 & -4.61 & 0 & 6211024 & 0.304 & 737387 & 16.96 & -- \\
ariane133 & SA & -- & Yes & -76.52 & -579878 & 80223 & 7135797 & 0.286 & 718246 & 0.240 & -- \\
ariane133 & CWI & 1/5 & Yes & -0.02 & -0.52 & 0 & 6105080 & 0.266 & 733327 & 814.4 & 329.4 \\
ariane133 & CWS & 5/5 & Yes & -0.02 & -0.56 & 0 & 5997840 & 0.265 & 732096 & 820.3 & 314.1 \\
ariane133 & PlaceReasoner-Beta & 3/5 & Yes & -0.02 & -0.35 & 0 & 5622394 & 0.258 & 728851 & 828.7 & 348.6 \\
ariane133 & Qwen3-VL-8B & -- & Fail: DRT-0073 & -- & -- & -- & -- & -- & -- & 3123.4 & 689.2 \\
ariane133 & Qwen3-VL-30B & 2/5 & Yes & -0.02 & -0.13 & 18166 & 6123046 & 0.283 & 728261 & 255.0 & 572.7 \\
ariane81 & RTL-MP & 2/5 & Yes & -0.14 & -8.43 & 0 & 5577716 & 0.184 & 694809 & 91.8 & -- \\
ariane81 & DREAMPlace & 1/5 & Fail: timeout & -- & -- & -- & -- & -- & -- & 9.53 & -- \\
ariane81 & ChiPFormer & 5/5 & Yes & -1.02 & -2103 & 0 & 6948313 & 0.177 & 700271 & 15.12 & -- \\
ariane81 & SA & -- & Fail: placed-only & -- & -- & -- & -- & -- & -- & 0.178 & -- \\
ariane81 & CWI & 1/5 & Yes & 0.00 & 0.00 & 0 & 4987790 & 0.171 & 693096 & 829.7 & 349.7 \\
ariane81 & CWS & 1/5 & Yes & 0.00 & 0.00 & 0 & 4987790 & 0.171 & 693096 & 836.1 & 337.2 \\
ariane81 & PlaceReasoner-Beta & 5/5 & Yes & -0.04 & -1.20 & 0 & 5203922 & 0.171 & 692142 & 847.6 & 362.6 \\
ariane81 & Qwen3-VL-8B & 2/5 & Yes & -0.09 & -6.24 & 0 & 6631628 & 0.198 & 693184 & 2449.0 & 759.3 \\
ariane81 & Qwen3-VL-30B & 1/5 & Yes & -0.09 & -13.95 & 0 & 6511947 & 0.203 & 694924 & 235.3 & 602.4 \\
bp\_be & RTL-MP & 4/5 & Yes & -0.28 & -23.54 & 0 & 2354232 & 0.123 & 240519 & 85.9 & -- \\
bp\_be & DREAMPlace & 4/5 & Yes & -0.47 & -29.81 & 63393 & 3814816 & -- & 243381 & 5.94 & -- \\
bp\_be & ChiPFormer & 1/5 & Yes & -0.58 & -54 & 0 & 2929620 & 0.131 & 243411 & 5.75 & -- \\
bp\_be & SA & -- & Yes & -0.24 & -21.26 & 37695 & 2411088 & 0.123 & 239997 & 0.032 & -- \\
bp\_be & CWI & 3/5 & Yes & -0.26 & -22.75 & 0 & 2598380 & 0.125 & 241666 & 693.7 & 198.5 \\
bp\_be & CWS & 3/5 & Yes & -0.26 & -22.75 & 0 & 2598379 & 0.125 & 241666 & 704.5 & 183.7 \\
bp\_be & PlaceReasoner-Beta & 1/5 & Yes & -0.13 & -9.42 & 0 & 2287187 & 0.118 & 238089 & 722.3 & 208.4 \\
bp\_be & Qwen3-VL-8B & 3/5 & Yes & -0.29 & -28.59 & 0 & 2435295 & 0.107 & 238967 & 349.7 & 308.9 \\
bp\_be & Qwen3-VL-30B & 1/5 & Yes & -0.16 & -13.28 & 0 & 2368477 & 0.102 & 238576 & 63.1 & 253.5 \\
vga\_lcd & RTL-MP & 5/5 & Yes & -0.78 & -498.02 & 45 & 2471807 & 0.182 & 650863 & 48.0 & -- \\
vga\_lcd & DREAMPlace & 3/5 & Yes & -0.85 & -146.70 & 62598 & 3804114 & -- & 650838 & 8.56 & -- \\
vga\_lcd & ChiPFormer & 2/5 & Yes & -0.82 & -1180 & 0 & 2880834 & 0.198 & 655252 & 11.15 & -- \\
vga\_lcd & SA & -- & Yes & -0.81 & -36.16 & 67293 & 2391472 & 0.184 & 649634 & 0.104 & -- \\
vga\_lcd & CWI & 3/5 & Yes & -0.66 & -48.84 & 0 & 2604281 & 0.184 & 650468 & 762.1 & 371.7 \\
vga\_lcd & CWS & 4/5 & Yes & -0.66 & -48.84 & 0 & 2604281 & 0.184 & 650468 & 774.9 & 364.1 \\
vga\_lcd & PlaceReasoner-Beta & 2/5 & Yes & -0.74 & -37.73 & 0 & 2342025 & 0.18 & 651477 & 785.1 & 386.4 \\
vga\_lcd & Qwen3-VL-8B & 4/5 & Yes & -0.89 & -41.92 & 0 & 2449050 & 0.190 & 648133 & 2084.3 & 730.6 \\
vga\_lcd & Qwen3-VL-30B & 2/5 & Yes & -0.71 & -27.86 & 0 & 2455068 & 0.192 & 647852 & 170.9 & 466.3 \\
\bottomrule
\end{tabular}
\vskip -12pt
\end{table*}

\begin{table*}[t]
\centering
\scriptsize
\setlength{\tabcolsep}{3pt}
\caption{Post-route results for four representative designs at the elongated ($2{:}1$)
aspect ratio; the remaining four designs are reported in the appendix.
``--'' denotes an unavailable metric and failure states indicate that detailed
routing did not complete. Winner \# reports the one-based position $k/N$ of the
selected candidate among the $N=5$ trials of that run.
Runtime and Token Usage are the summed model-driver time and the total of model
input and output tokens, and are reported for the model-driven flows
(\textsc{CWI}, \textsc{CWS}, \textsc{PlaceReasoner}, and the two Qwen
baselines); the non-LLM baselines incur no model cost.}
\vskip -6pt
\label{tab:pr_detail_ar2_4}
\begin{tabular}{llcrrrrrrrcc}
\toprule
Design & Method & Winner \# & Route & WNS (ns) & TNS (ns) & DRC & WL ($\mu$m) & Power (W) & Area ($\mu$m$^2$) & Runtime (s) & Token Usage (k) \\
\midrule
ariane133 & RTL-MP & 1/5 & Yes & -0.12 & -25.29 & 0 & 6033112 & 0.276 & 731337 & -- & -- \\
ariane133 & DREAMPlace & -- & Fail: timeout & -- & -- & -- & -- & -- & -- & -- & -- \\
ariane133 & ChiPFormer & 2/5 & Yes & -0.39 & -620 & 0 & 7079270 & 0.278 & 738872 & -- & -- \\
ariane133 & SA & -- & Fail: GP1 & -- & -- & -- & -- & -- & -- & -- & -- \\
ariane133 & \textbf{PlaceReasoner} & 4/5 & Yes & -0.02 & -0.08 & 0 & 5690597 & 0.285 & 730228 & 834.9 & 357.1 \\
ariane133 & Qwen3-VL-8B & -- & Fail: GRT-0116 & -- & -- & -- & -- & -- & -- & 2228.3 & 479.9 \\
ariane133 & Qwen3-VL-30B & -- & Fail: DRT-0073 & -- & -- & -- & -- & -- & -- & 208.9 & 600.6 \\
ariane81 & RTL-MP & 2/5 & Yes & -0.10 & -35.10 & 0 & 6180469 & 0.179 & 696239 & -- & -- \\
ariane81 & DREAMPlace & -- & Fail: timeout & -- & -- & -- & -- & -- & -- & -- & -- \\
ariane81 & ChiPFormer & 4/5 & Yes & -1.82 & -5617 & 0 & 9503525 & 0.181 & 727526 & -- & -- \\
ariane81 & SA & -- & Fail: capped & -- & -- & -- & -- & -- & -- & -- & -- \\
ariane81 & \textbf{PlaceReasoner} & 1/5 & Yes & -0.02 & -0.15 & 0 & 5798709 & 0.194 & 698754 & 861.2 & 401.7 \\
ariane81 & Qwen3-VL-8B & 1/5 & Yes & -0.40 & -310.10 & 28 & 7017977 & 0.202 & 694691 & 778.8 & 330.1 \\
ariane81 & Qwen3-VL-30B & 2/5 & Yes & -0.09 & -43.12 & 1 & 5923522 & 0.193 & 692020 & 524.7 & 579.2 \\
bp\_be & RTL-MP & 1/5 & Yes & -0.37 & -38.96 & 0 & 2652279 & 0.123 & 240762 & -- & -- \\
bp\_be & DREAMPlace & -- & Fail: timeout & -- & -- & -- & -- & -- & -- & -- & -- \\
bp\_be & ChiPFormer & -- & Fail: segfault & -- & -- & -- & -- & -- & -- & -- & -- \\
bp\_be & SA & -- & Yes & -0.36 & -36.62 & 34190 & 2749978 & 0.124 & 241710 & -- & -- \\
bp\_be & \textbf{PlaceReasoner} & 4/5 & Yes & -0.22 & -18.06 & 1 & 2873409 & 0.109 & 240882 & 728.9 & 241.6 \\
bp\_be & Qwen3-VL-8B & 1/5 & Yes & -0.39 & -31.03 & 0 & 2647452 & 0.107 & 239686 & 216.1 & 237.2 \\
bp\_be & Qwen3-VL-30B & 1/5 & Fail: GRT-0116 & -- & -- & -- & -- & -- & -- & 256.2 & 480.3 \\

\bottomrule
\end{tabular}
\vskip -6pt
\end{table*}

% \subsubsection{Results on PlaceReasoner-Bench.}

\subsection{Post-Route Placement Quality: Different Aspect Ratios and Analysis}

Tables~\ref{tab:pr_detail_ar1_4} and~\ref{tab:pr_detail_ar2_4} report routed WNS, TNS, DRC violations, wirelength, power, and area for four representative designs at $1{:}1$ and $2{:}1$ aspect ratios, respectively. 
Complete results for all eight designs are provided in the appendix (Tables~\ref{tab:pr_detail_ar1_full} and~\ref{tab:pr_detail_ar2}). 
We compare PlaceReasoner-Beta with four classical baselines; the \textsc{CWI} and \textsc{CWS} ablations are discussed separately.
Three findings stand out. 
\textbf{(i) Better PPA:} Among DRC-clean runs, PlaceReasoner-Beta achieves the best timing on all eight $1{:}1$ tasks and six of eight $2{:}1$ tasks, without sacrificing power or area. \textbf{(ii) Better wirelength:} It also reduces routed wirelength on most tasks, despite never explicitly optimizing wirelength, suggesting that geometric reasoning can improve downstream routing quality beyond the proxy objective. \textbf{(iii) Better routability:} PlaceReasoner-Beta is the only method that produces a routable layout with at most two DRC violations on all 16 tasks. 
We first examine the $1{:}1$ tasks in detail, then highlight how performance changes under the more challenging $2{:}1$ geometry, and finally analyze why the reasoning-driven approach produces these gains.

\iffalse

Tables~\ref{tab:pr_detail_ar1_4} and~\ref{tab:pr_detail_ar2_4} report routed WNS, TNS,
DRC violations, wirelength, power, and area at the two aspect ratios for four
representative designs.
Due to the page limit, the complete results over all eight designs are given in the appendix
(Tables~\ref{tab:pr_detail_ar1_full} and~\ref{tab:pr_detail_ar2}).
This section compares PlaceReasoner-Beta against the four classical baselines; the
\textsc{CWI} and \textsc{CWS} rows of the same tables are analyzed separately in
the ablation study.
Three observations organize the discussion.
\textbf{(i)~Better PPA:} among the DRC-clean baseline runs, PlaceReasoner-Beta attains
the best timing on all eight square designs and on six of eight elongated ones,
and it does so at equal or lower power and area rather than by trading them away.
\textbf{(ii)~Better wirelength, unexpectedly:} routed wirelength shrinks on most
square designs even though wirelength never enters the objective, the checkers,
or the reward.
\textbf{(iii)~Better DRC:} PlaceReasoner-Beta is the only method that produces a
routable, essentially violation-free layout on all $16$ tasks.
We treat the square ($1{:}1$) tasks as the detailed example to show these findings, then report only
where the elongated ($2{:}1$) tasks differ, and finally explain the underlying
mechanism that distinguishes PlaceReasoner-Beta from others.

\fi

% \textbf{Square ($1{:}1$) tasks.}

\subsubsection{Square ($1{:}1$) Tasks} 
\begin{figure}[t]
\centering
% \vskip -15pt
\includegraphics[width=1.0\linewidth]{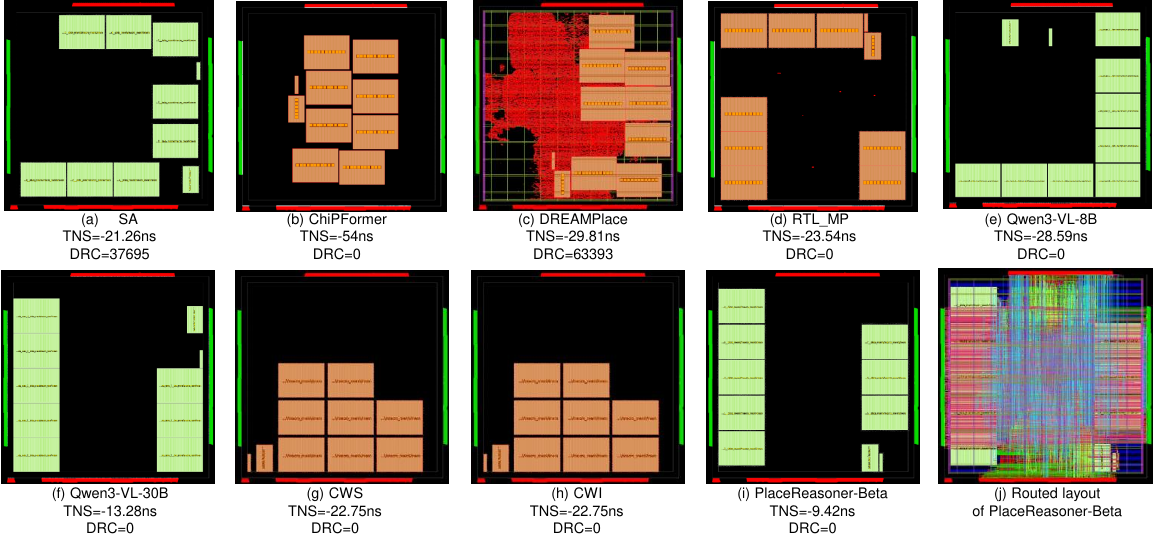}
\vskip -6pt
\caption{Macro placements produced by different methods on \texttt{bp\_be} at
aspect ratio $1{:}1$, each annotated with the post-route TNS and DRC count it
leads to; (j) is the routed layout obtained from the placement in (i).
(a) SA and (c) DREAMPlace spread macros over the core interior and leave narrow,
irregular gaps between them and the die boundary; the router cannot absorb the
standard cells there, yielding $37{,}695$ and $63{,}393$ violations (in (c), red
marks the congested region).
(b) ChiPFormer and (d) RTL-MP route DRC-clean, but their interior macro columns
cut the standard-cell area into disconnected strips, and timing stalls at
$-54$ and $-23.54$~ns.
(i) PlaceReasoner-Beta instead pushes every macro against the periphery in a
single contiguous column, preserving one large standard-cell region and clear
routing channels, and reaches $-9.42$~ns with zero violations.
The same effect is visible across backbones: (f) Qwen3-VL-30B also drives the
macros to the periphery and reaches $-13.28$~ns, whereas (e) Qwen3-VL-8B leaves
them fragmented in the interior and stalls at $-28.59$~ns.
Panels (g) and (h) are the \textsc{CWS} and \textsc{CWI} ablations, discussed in
the ablation study.}
\label{fig:layout}
\vskip -6pt
\end{figure}

\textbf{PPA are superior.}
Table~\ref{tab:pr_detail_ar1_4} shows that \textbf{PlaceReasoner-Beta achieves the best WNS and TNS among all DRC-clean baseline runs on all eight designs}.
The advantage is largest when macro placement strongly constrains the design.
On \texttt{ariane133}, it achieves $-0.02$/$-0.35$~ns versus $-0.10$/$-4.60$~ns for RTL-MP and $-0.04$/$-4.61$~ns for ChiPFormer, a $92.4\%$ TNS reduction over the strongest baseline. 
On \texttt{vga\_lcd}, it reduces TNS from $-498.02$~ns (RTL-MP) to $-37.73$~ns.
We use \texttt{bp\_be} as an exemplary case to illustrate how placement drives downstream quality (Fig.~\ref{fig:layout}).
Specifically, DREAMPlace and SA complete routing but produce $63{,}393$ and $37{,}695$ DRC violations, respectively, making their layouts physically invalid.
Among DRC-clean baselines, RTL-MP and ChiPFormer achieve WNS/TNS of $-0.28$/$-23.54$~ns and $-0.58$/$-54.00$~ns, whereas \textbf{PlaceReasoner-Beta} improves them to \textbf{$-0.13$/$-9.42$}~ns, reducing WNS/TNS deficits by $53.6\%$/$60.0\%$ over RTL-MP and $77.6\%$/$82.6\%$ over ChiPFormer.
The invalid layouts are visibly congested: DREAMPlace concentrates macros in the right half of the core (Fig.~\ref{fig:layout}(c)), while ChiPFormer (Fig.~\ref{fig:layout}(b)) and RTL-MP (Fig.~\ref{fig:layout}(d)) place them internally, fragmenting the standard-cell region. 
In contrast, \textbf{PlaceReasoner-Beta} arranges macros along the die edges (Fig.~\ref{fig:layout}(i)), preserving a contiguous standard-cell region and open routing channels; the resulting routed layout is shown in Fig.~\ref{fig:layout}(j).

\begin{figure*}[!t]
  \centering
  \begin{minipage}[b]{0.49\linewidth}
    \centering
    \includegraphics[width=\linewidth]{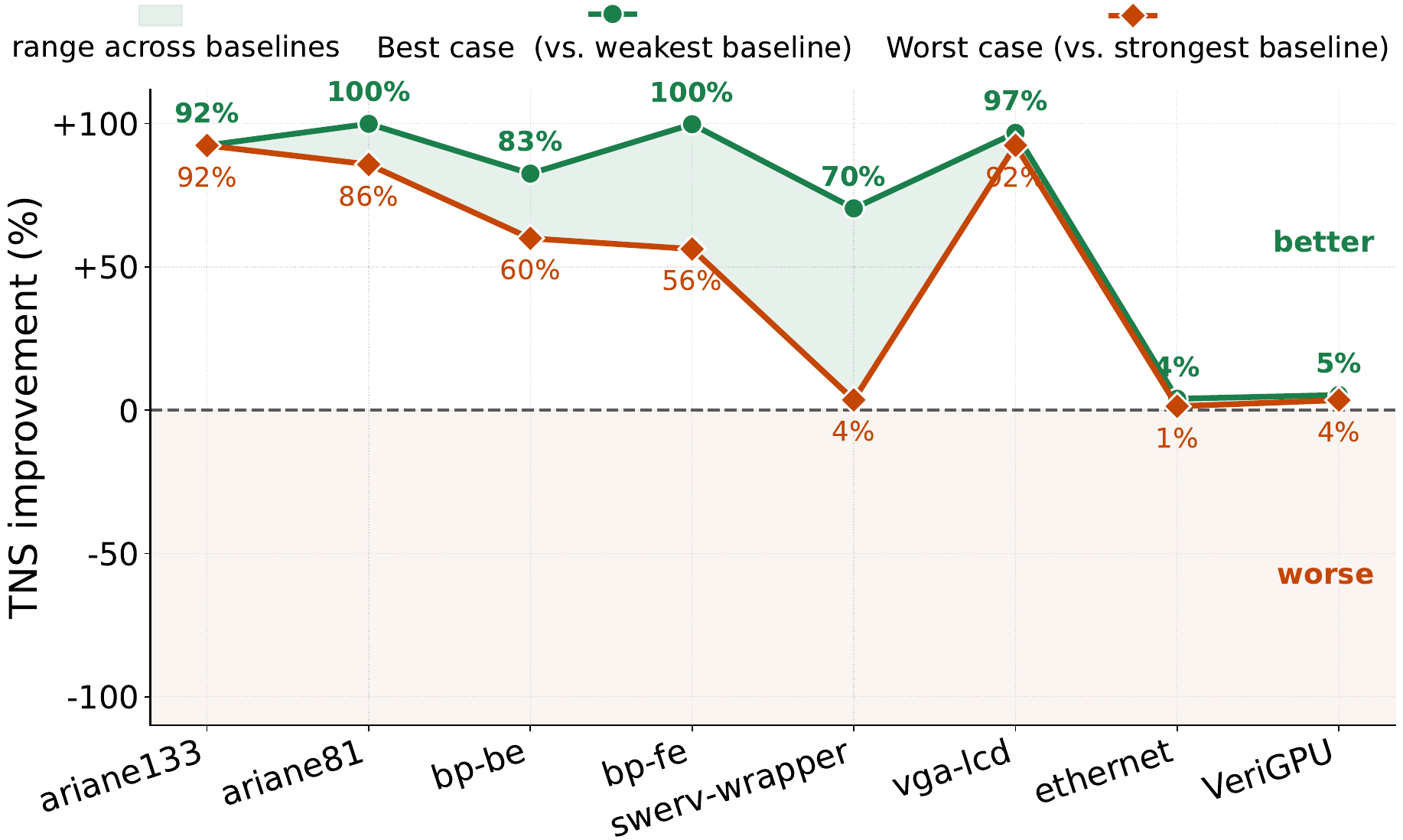}
  \end{minipage}\hfill
  \begin{minipage}[b]{0.49\linewidth}
    \centering
    \includegraphics[width=\linewidth]{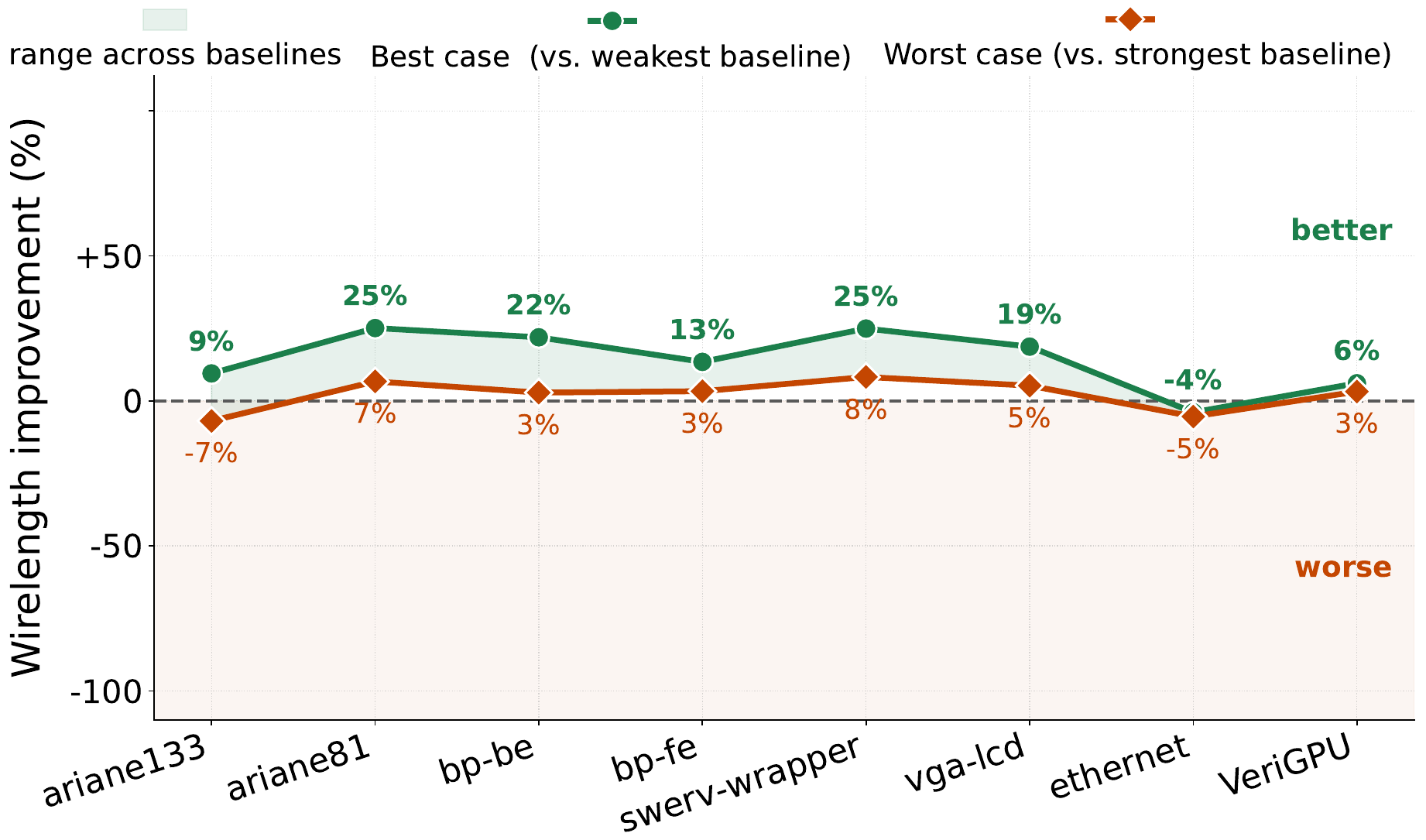}
  \end{minipage}
  \vskip -6pt
  \caption{Performance improvement of PlaceReasoner-Beta on post-route quality on the
  square ($1{:}1$) tasks: best- vs.\ worst-case improvement over the baselines,
  per design. \textbf{Left:}  TNS; \textbf{right:} routed
  wirelength. For each design the \emph{best case} (green) is the improvement
  over the weakest baseline and the \emph{worst case} (orange) over the
  strongest; the shaded band spans the two. Improvement is
  $(\lvert x_{\text{base}}\rvert-\lvert x_{\textsc{pr}}\rvert)/
  \max(\lvert x_{\text{base}}\rvert,\lvert x_{\textsc{pr}}\rvert)$.
  That is, $+100\%$ means \text{PlaceReasoner-Beta} drives the metric to zero and positive
  values (green zone) mean it is better. Although \text{PlaceReasoner-Beta}
  never optimizes wirelength, it still shortens it over the baselines on most
  designs.}
  \vskip -6pt
  \label{fig:pr_improvement}
\end{figure*}

Across all eight designs, PlaceReasoner-Beta reduces TNS by \textbf{$57.6\%$ on average relative to the strongest baseline per design}, and by \textbf{$61.2\%$ relative to the four-baseline field} when counting only runs within the $1{,}000$-violation DRC budget (Fig.~\ref{fig:pr_improvement}, left). 
The first metric measures improvement over the best available competitor, while the second captures improvement over the overall baseline field. 
Total power is lower than RTL-MP on six designs and equal on \texttt{ethernet}, with only a minor regression on \texttt{swerv\_wrapper} ($0.243$ vs.\ $0.239$~W, $+1.7\%$). 
Post-route area is also the smallest among DRC-clean baselines on all eight designs, e.g., $238{,}089$~$\mu$m$^2$ on \texttt{bp\_be} versus $240{,}519$~$\mu$m$^2$ for RTL-MP and $243{,}411$~$\mu$m$^2$ for ChiPFormer. 
Gains are smaller on \texttt{VeriGPU} and \texttt{ethernet}, where macros occupy only $30.5\%$ and $14.2\%$ of core area, leaving downstream optimization more freedom to recover from mediocre macro placement.

\noindent \textbf{Wirelength improves despite never being optimized.}
Although PlaceReasoner-Beta never optimizes wirelength, its routed wirelength is shorter than RTL-MP on six of eight designs (up to $8.2\%$ on \texttt{swerv\_wrapper} and $6.7\%$ on \texttt{ariane81}) and shorter than ChiPFormer on seven of eight (up to $25.1\%$ on \texttt{ariane81}) (Fig.~\ref{fig:pr_improvement}, right). 
The contrast with DREAMPlace is particularly striking: despite directly minimizing an analytical wirelength objective, its routed wirelength is $37$--$97\%$ longer on the five completed designs, e.g., $3.81\times10^{6}$ versus $2.29\times10^{6}$~$\mu$m on \texttt{bp\_be}. 
Its wirelength-optimal macro positions can force routing detours and via stacks that pre-route objectives do not capture. 
\textbf{Thus, optimizing a wirelength proxy does not necessarily improve routed wirelength; respecting routability and dataflow structure can improve it as a by-product.}

\noindent\textbf{DRC \& routability.}
PlaceReasoner-Beta is DRC-clean on all $1{:}1$ designs.
RTL-MP leaves 14 and 45 violations on \texttt{ethernet} and \texttt{vga\_lcd}; SA is clean only on \texttt{bp\_fe}, leaves $27{,}609$--$80{,}223$ violations elsewhere, and fails on \texttt{ariane81}. 
DREAMPlace times out on three designs and leaves $30{,}699$--$65{,}340$ violations on the five it completes. 
ChiPFormer matches our DRC record, but at substantial timing cost, with TNS reaching $-2103$~ns on \texttt{ariane81} and $-1384$~ns on \texttt{swerv\_wrapper}. 
That is, it achieves legality through conservative placements that do not adequately preserve timing.

\subsubsection{Elongated ($2{:}1$) Tasks}
\begin{figure}[t]
  \centering
  \begin{minipage}[b]{0.49\linewidth}
    \centering
    \includegraphics[width=\linewidth]{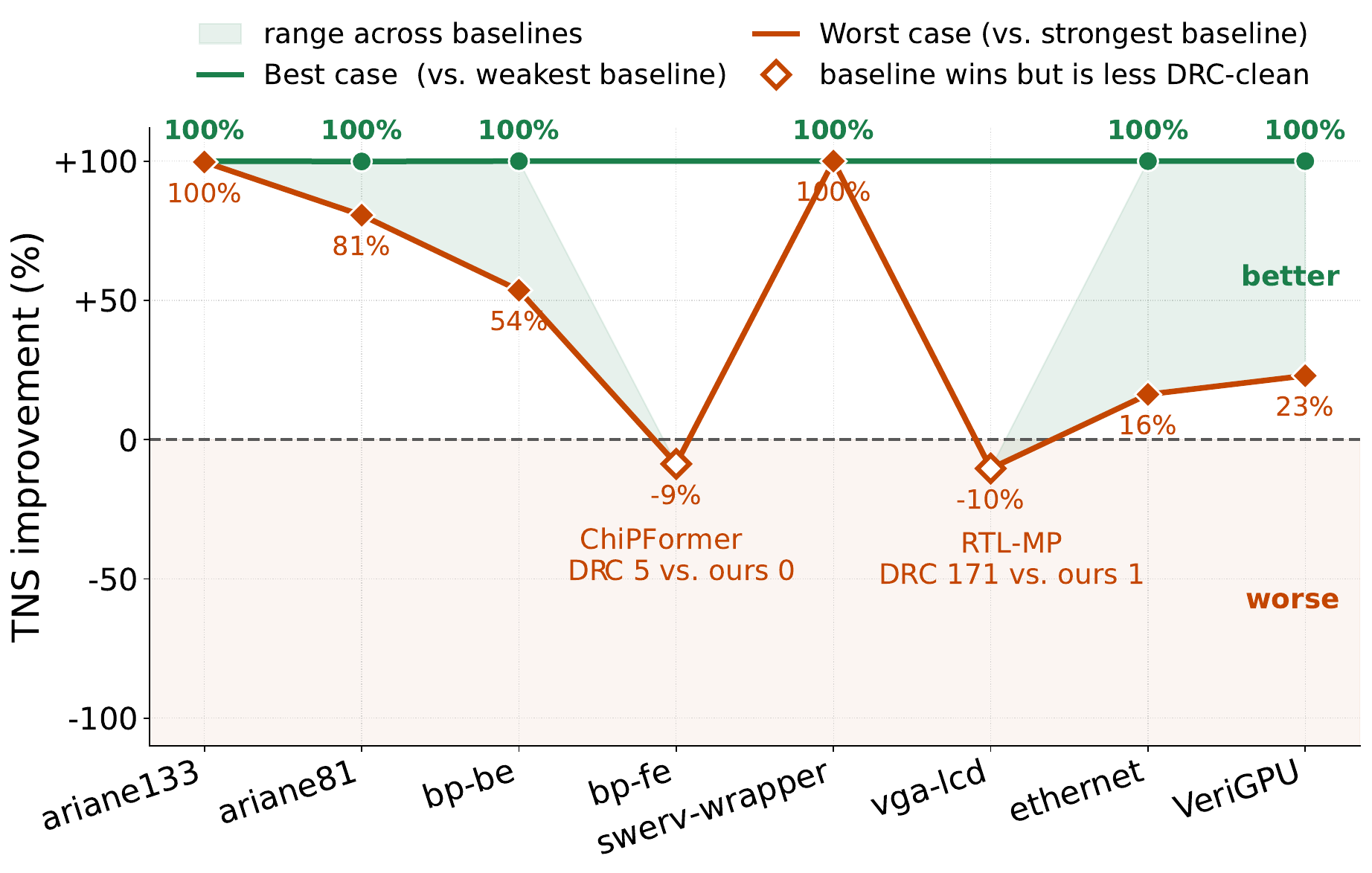}
  \end{minipage}\hfill
  \begin{minipage}[b]{0.49\linewidth}
    \centering
    \includegraphics[width=\linewidth]{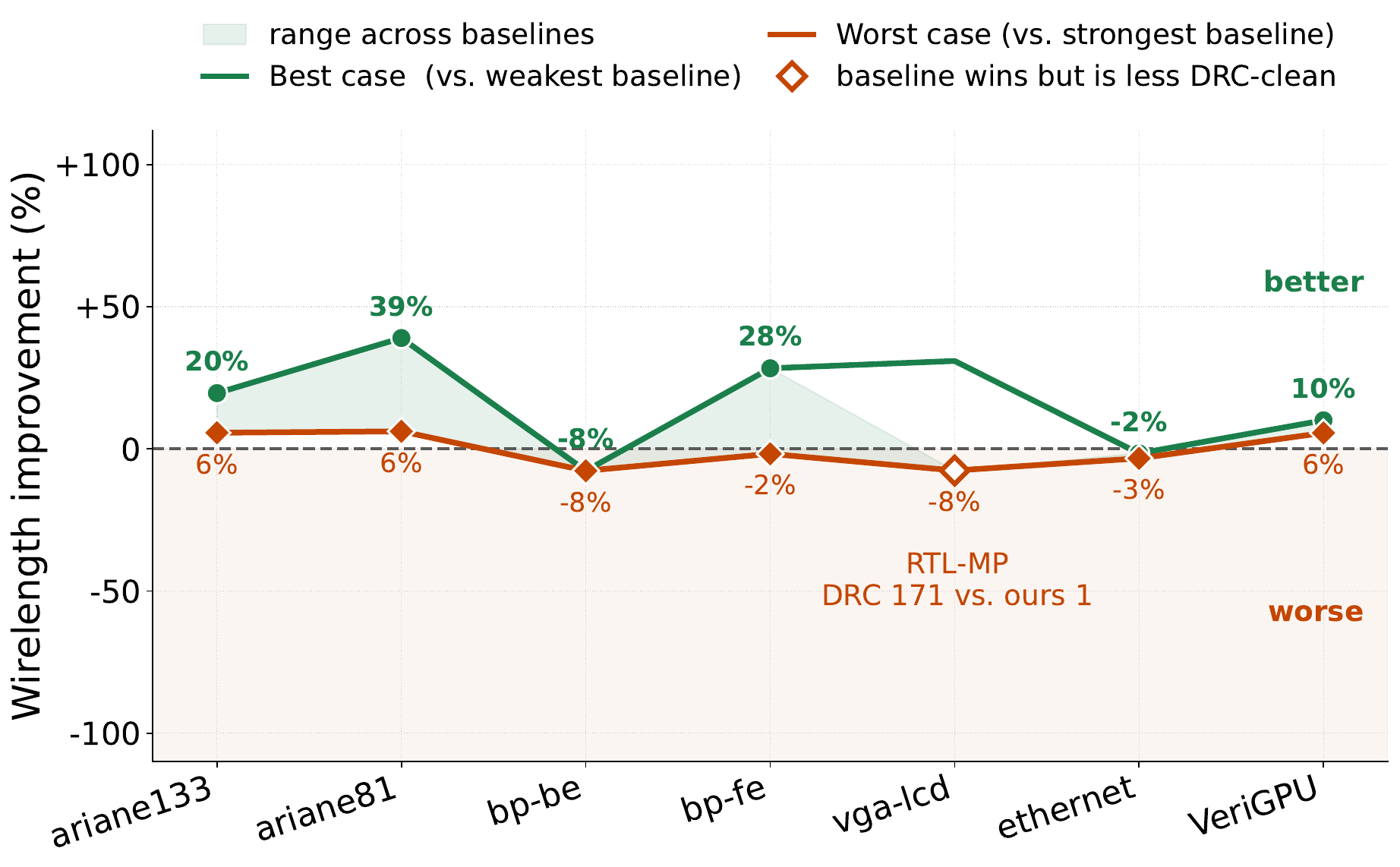}
  \end{minipage}
    \vskip -6pt
  \caption{Post-route improvement of PlaceReasoner-Beta on the elongated ($2{:}1$)
  tasks: best- vs.\ worst-case improvement over the four classical baselines,  per design. \textbf{Left:} TNS; \textbf{right:} routed wirelength. For each
  design, the \emph{best case} (green) is the improvement over the weakest
  baseline and the \emph{worst case} (orange) over the strongest; the shaded
  band spans the two. Improvement is
  $(\lvert x_{\text{base}}\rvert-\lvert x_{\textsc{pr}}\rvert)/
  \max(\lvert x_{\text{base}}\rvert,\lvert x_{\textsc{pr}}\rvert)$, so $+100\%$
  means the metric is driven to zero and positive values (green zone) mean
  PlaceReasoner-Beta is better. A baseline that fails to route, or routes with more
  than $10{,}000$ DRC violations, counts as a failed run: for TNS it scores a
  full $+100\%$ win, which is why the best-case curve saturates; for wirelength
  it is excluded instead, and \texttt{swerv\_wrapper} therefore drops from that
  panel because no baseline survives. Hollow markers on \texttt{bp\_fe} and
  \texttt{vga\_lcd} mark the two designs where the strongest baseline attains
  marginally better TNS but with more DRC violations than our run. Timing
  therefore improves on six of the eight designs, with a mean worst-case gain of
  $+44.3\%$, while routed wirelength is on par with the strongest baseline.}
    \vskip -6pt
  \label{fig:pr_improvement_ar2}
\end{figure}

Table~\ref{tab:pr_detail_ar2_4} shows that the elongated core is substantially harder: no classical baseline routes all eight designs with fewer than 10 DRC violations. Detailed routing fails outright on five of eight designs for DREAMPlace, four for SA,
and one for ChiPFormer, and several completed runs remain DRC-invalid, with up to
$62{,}832$ violations.
\textbf{PlaceReasoner-Beta is the only method that routes all eight designs with at most two DRC violations}, with five completely DRC-clean and the remaining three having only one or two violations. 
In contrast, the worst cases reach $1{,}421$, $1{,}456$, and $62{,}832$ violations for RTL-MP, ChiPFormer, and DREAMPlace, respectively.
Among baseline runs that complete within the $1{,}000$-violation DRC budget, \textbf{PlaceReasoner-Beta achieves the best TNS on six of eight designs}, reducing TNS by $44.3\%$ relative to the strongest surviving baseline and $53.0\%$ relative to the four-baseline field {(Fig.~\ref{fig:pr_improvement_ar2})}. 
On the two exceptions, the differences are small but come with worse routability: ChiPFormer achieves $-0.21$~ns on \texttt{bp\_fe} versus $-0.23$~ns for PlaceReasoner-Beta but with five DRC violations, while RTL-MP reaches $-29.70$~ns on \texttt{vga\_lcd} versus $-33.13$~ns with 171 violations, compared with only one for PlaceReasoner-Beta.

\begin{figure}[t]
\centering
% \vskip -15pt
\includegraphics[width=1.0\linewidth]{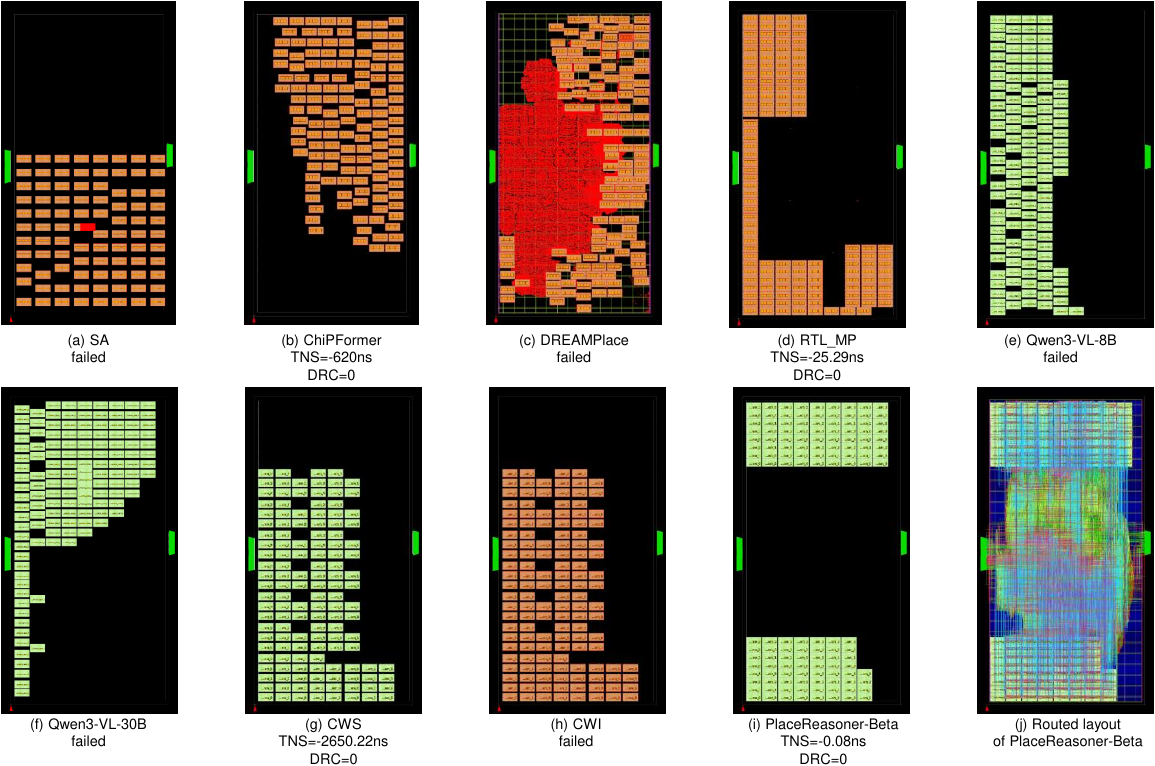}
\vskip -6pt
\caption{Macro placements on \texttt{ariane133} at aspect ratio $2{:}1$, annotated
with the post-route TNS and DRC count each leads to; (j) is the routed layout of
(i), and ``failed'' marks runs whose detailed routing did not complete.
The elongated core is unforgiving: five of the nine placements never route.
(a), (c) and (f) leave the macros in a dense interior cluster (in (c), red marks
the congested region), while (b) and (e) fragment the standard-cell area into
disconnected strips; only ChiPFormer routes, at $-620$~ns.
(d) RTL-MP is the strongest baseline, holding an L-shaped ring along the left
and bottom edges and reaching $-25.29$~ns DRC-clean.
(i) \textbf{PlaceReasoner-Beta} instead packs the macros into two bands against
the short peripheries, leaving one uninterrupted standard-cell region across the
die, and reaches $-0.08$~ns with zero violations.
Panels (g) and (h) are the \textsc{CWS} and \textsc{CWI} ablations, discussed in
the ablation study.}
\label{fig:ariane133}
\vskip -6pt
\end{figure}

The key difference between the two aspect ratios is \textbf{reliability rather than simply timing margin}. 
Failed routing runs increase from $4/32$ at $1{:}1$ to $10/32$ at $2{:}1$, while even RTL-MP drops from six to four DRC-clean designs. 
PlaceReasoner-Beta, in contrast, routes all eight designs at both aspect ratios with at most two violations. 
Thus, under elongated geometries, the primary benefit of explicit spatial reasoning is improved legality and routability, the properties most stressed when the floorplan becomes asymmetric.
We use \texttt{ariane133} as an exemplary case to show this advantage (Fig.~\ref{fig:ariane133}). 
DREAMPlace times out and SA fails during global placement, while RTL-MP and ChiPFormer complete routing with WNS/TNS of $-0.12$/$-25.29$~ns and $-0.39$/$-620$~ns, respectively. \textbf{PlaceReasoner-Beta} improves these to $-0.02$/$-0.08$~ns, a $99.7\%$ TNS reduction over RTL-MP, while also reducing routed wirelength by $5.7\%$ and $19.6\%$ relative to the two baselines.
The layouts reveal why: SA (Fig.~\ref{fig:ariane133}(a)) and DREAMPlace (Fig.~\ref{fig:ariane133}(c)) leave macros clustered in the interior, creating severe congestion, while ChiPFormer (Fig.~\ref{fig:ariane133}(b)) scatters them across the upper core and suffers extreme timing degradation.
RTL-MP places most macros along the top and left edges (Fig.~\ref{fig:ariane133}(d)), but only \textbf{PlaceReasoner-Beta} organizes them into two bands along the short peripheries (Fig.~\ref{fig:ariane133}(i)), preserving a continuous standard-cell region across the die (see the routed result shown in Fig.~\ref{fig:ariane133}(j)).

As at $1{:}1$, gains are smaller for macro-sparse designs such as \texttt{VeriGPU} and \texttt{ethernet}, with $23.0\%$ and $17.3\%$ TNS reductions over RTL-MP. 
Unlike the square tasks, however, $2{:}1$ provides no additional wirelength or power advantage: PlaceReasoner-Beta has shorter routed wirelength on three of eight designs and is essentially tied overall (geometric-mean ratio $0.998$), while power is lower on three. 
Since neither metric is directly optimized, the elongated geometry leaves less freedom to improve timing, routability, and other post-route objectives simultaneously.
\textbf{The key gain is therefore robust timing closure and routability, achieved without sacrificing overall wirelength or power.}

% \textbf{Why does it work?}

\subsubsection{Why Does PlaceReasoner-Beta Work Better?}
Figs.~\ref{fig:pr_improvement} and~\ref{fig:layout} suggest a common mechanism: PlaceReasoner-Beta produces macro arrangements that preserve the geometric structures favored by downstream physical design.
The analytical and annealing baselines tend to place macros in the core interior, creating narrow, irregular slivers of standard-cell area between macros and along the die boundary. 
These regions are difficult to fill and route, leading to congestion, detours, and ultimately the DRC violations reported in Tables~\ref{tab:pr_detail_ar1_4} and~\ref{tab:pr_detail_ar2_4}.
In contrast, PlaceReasoner-Beta places macros along the periphery, keeps related clusters contiguous with consistent orientations, and preserves a large, connected standard-cell region. 
This geometry naturally reduces routing bottlenecks, provides more flexibility for timing-critical cell placement, and shortens routing detours, explaining the observed gains in DRC, WNS/TNS, and, as a by-product, routed wirelength and power. 
These behaviors reflect established expert placement principles~\cite{macro_guide}, which the skill pack expresses as explicit reasoning rules rather than a scalar optimization term.
The VLM enables this reasoning by grounding decisions in the floorplan image. 
It can recognize which regions remain available, identify narrowing channels, detect fragmented clusters, and revise placements in response to checker feedback. 
In contrast, analytical placers must encode such structural preferences indirectly through differentiable objectives; as the DREAMPlace results illustrate, minimizing a wirelength proxy does not guarantee a geometrically or physically favorable routed design.

\subsection{Visual Grounding: Macro Orientation and Pin-Side Alignment}

\begin{figure}[!t]
\centering
\includegraphics[width=1.0\linewidth]{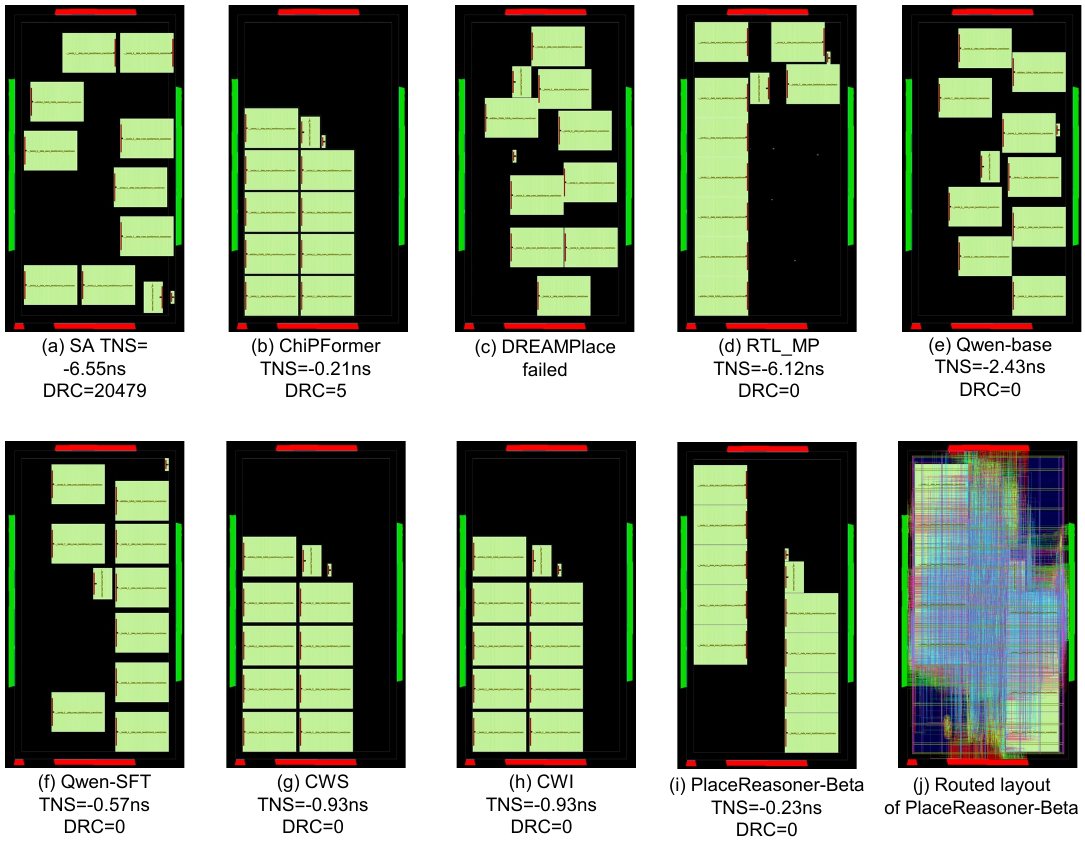}
  \vskip -6pt
\caption{Macro placements of \texttt{bp\_fe} at aspect ratio $2{:}1$, with each
macro's pin side marked by the dark bar on its edge; green and red bands are the
I/O pins on the die boundary, and (j) is the routed layout of (i).
Only (i) \textsc{PlaceReasoner-Beta} turns every pin side toward the standard-cell
region: it mirrors the orientation of the two macro groups so that the left group
faces right and the right group faces left, both onto the central corridor.
(a) SA orients macros inconsistently across a scattered interior placement
($20{,}479$ violations); (b) ChiPFormer packs one abutting block with uniformly
oriented columns, so the left column faces the die boundary and the right column
faces the back of the left, and it leaves $5$ violations; (d) RTL-MP orients its
main column inward but leaves the top-right macro facing the boundary beside a
large dead region; (c) DREAMPlace does not complete routing.
(e)--(f) \textsc{Qwen3-VL-8B} and \textsc{Qwen3-VL-30B}, and (g)--(h) the two
ablations, apply a single orientation to every macro, so their left-hand macros
face outward.}
  \vskip -6pt
\label{fig:bp_fe_pin_side}
\end{figure}

Macro placement is defined by both position and orientation.
A key expert principle is to orient each macro so that its pin side faces the standard-cell logic it communicates with, rather than a die boundary or the back of a neighboring macro~\cite{macro_guide}.
This is where reasoning over layout geometry can outperform objective-driven optimization: flipping a macro may leave wirelength and density objectives nearly unchanged, yet determine whether its pins face usable routing resources or a physical obstruction.
Fig.~\ref{fig:bp_fe_pin_side} shows this distinction by using \texttt{bp\_fe} at $2{:}1$ as an example, with each macro's pin side highlighted.
\textbf{PlaceReasoner-Beta is the only method that correctly aligns the pin sides of all macros}. 
It divides the macros into left and right groups and mirrors their orientations, directing all pins toward the central standard-cell corridor. 
In contrast, SA scatters macros through the core with inconsistent orientations and produces $20{,}479$ DRC violations. 
ChiPFormer packs the macros into an abutting block with uniform orientations, causing pins on the outer columns to face the boundary or neighboring macros. 
RTL-MP orients the main column appropriately but leaves the top-right macro facing the boundary and creates a large dead region.
Even the Qwen3-VL backbones, despite producing sensible macro locations, assign a common orientation to the macros, suggesting that they capture \emph{where} macros belong more readily than \emph{which way} they should face.

\begin{figure}[!t]
\centering
\includegraphics[width=1.0\linewidth]{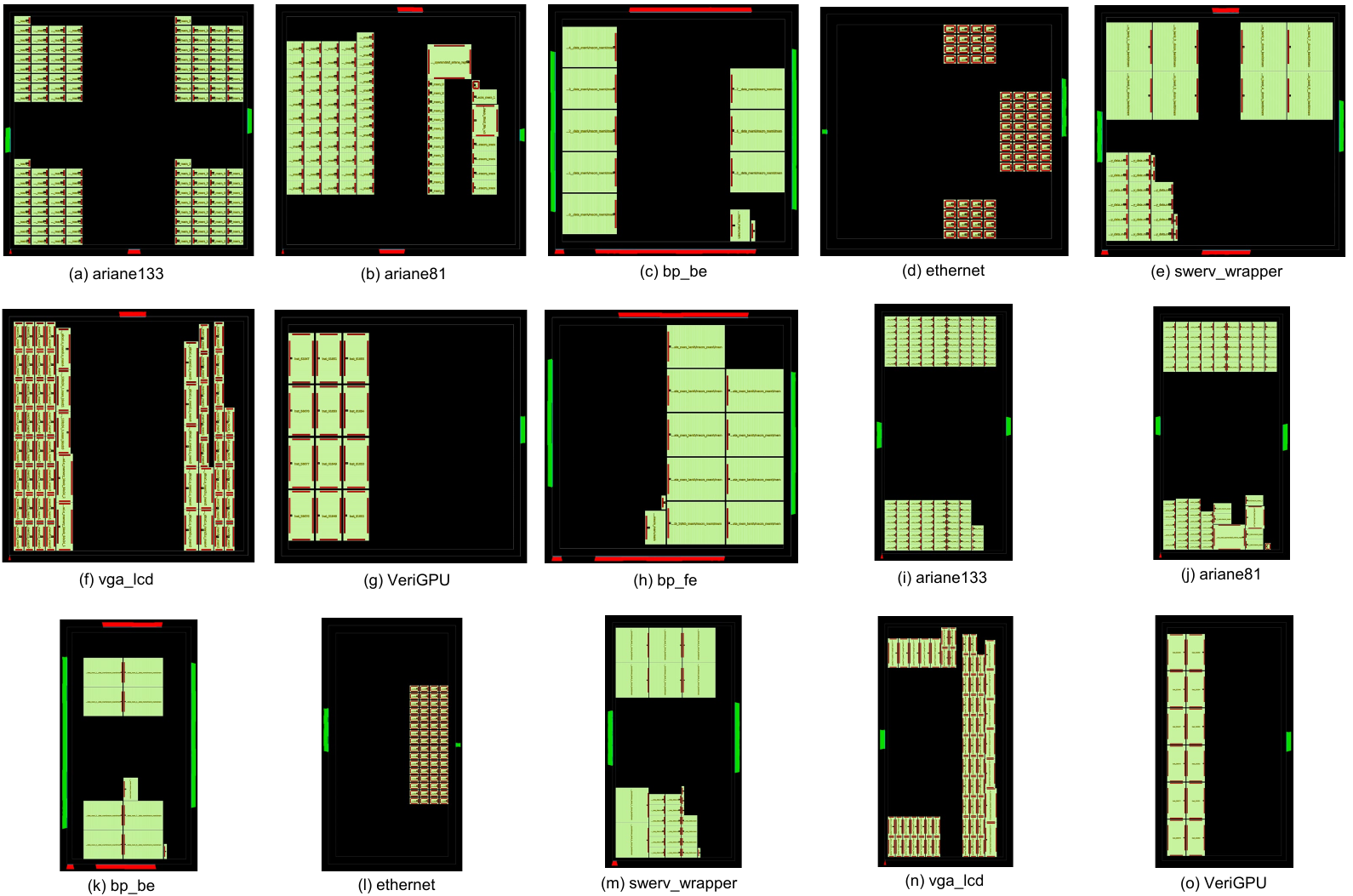}
  \vskip -6pt
\caption{Macro placements produced by PlaceReasoner-Beta on the remaining 15
PlaceReasoner-Bench tasks, with each macro's pin side marked:
(a)--(h) the eight designs at aspect ratio $1{:}1$ and (i)--(o) seven of them at
$2{:}1$; \texttt{bp\_fe} at $2{:}1$ is the case of Fig.~\ref{fig:bp_fe_pin_side}
and is not repeated here.
Across designs and aspect ratios, the macros are grouped against the periphery and
each group is oriented so that its pin side faces the standard-cell region it
serves, leaving that region contiguous.}
  \vskip -12pt
\label{fig:all_macro_placement}
\end{figure}

Correct orientation enables multiple physical objectives to be satisfied simultaneously. ChiPFormer achieves the best baseline timing ($-0.21$~ns WNS) but leaves five DRC violations and no clear routing channel; SA is highly DRC-invalid; and RTL-MP is DRC-clean but has $27\times$ worse TNS ($-6.12$~ns), partly due to its dead region and misoriented macro.
\textbf{PlaceReasoner-Beta achieves $-0.23$~ns WNS with zero violations}, remaining within $0.02$~ns of the best timing while preserving both routability and pin-side alignment. 
This example highlights the advantage of multimodal reasoning: \textbf{pin orientation, dataflow locality, and routing space can be evaluated jointly in the layout, whereas they are difficult to express through a single scalar objective}. 
Fig.~\ref{fig:all_macro_placement} shows the remaining 15 benchmark tasks, where the same pattern recurs: PlaceReasoner-Beta places macro groups near the periphery and orients their pin sides toward the standard-cell regions they serve.

\subsection{Exploration Robustness: Candidate Spread and PPA Variance}

PlaceReasoner-Beta deliberately explores multiple placement hypotheses rather than committing to a single solution. 
For each design, the planner generates $N=5$ structurally distinct candidates, which are screened by the checkers and post-route optimizer to select the best verified solution. 
Candidate variation therefore represents \emph{search breadth} rather than method instability: exploring alternative macro organizations is how the framework discovers high-quality layouts.
Tables~\ref{tab:variance_summary} and~\ref{tab:variance_by_cell} quantify this behavior across 80 candidate runs ($8$ designs $\times$ $2$ aspect ratios $\times$ $5$ candidates). 
Of these, 68 complete routing and 42 are DRC-clean. The variation is strongly metric-dependent. Timing varies substantially across candidates, with mean CVs of $80.4\%$ for TNS and $43.7\%$ for WNS, whereas power and area are nearly invariant, with mean CVs of only $1.56\%$ and $0.27\%$ and maxima of $5.3\%$ and $1.0\%$, respectively. 
\textbf{Thus, macro placement provides substantial leverage over timing but limited leverage over power and area}, motivating candidate generation and screening rather than one-shot placement.

\begin{figure}[t]
  \centering
  \includegraphics[width=1.0\linewidth]{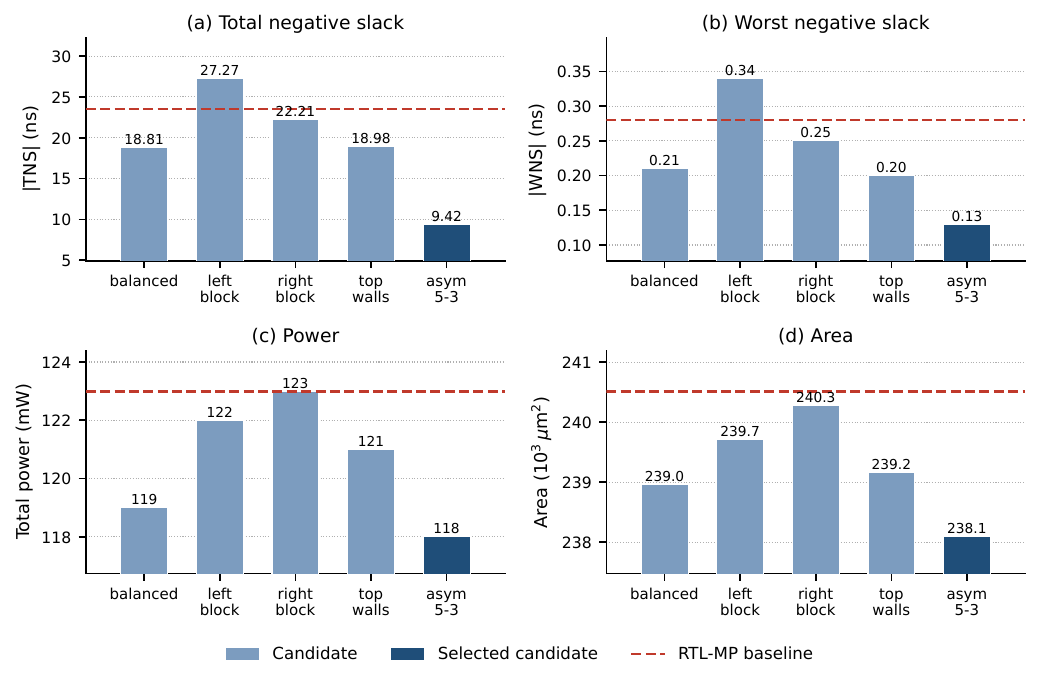}
    \vskip -6pt
  \caption{PPA of the five structurally distinct macro placements PlaceReasoner-Beta
  explores for \texttt{bp\_be} at aspect ratio $1{:}1$, against RTL-MP (dashed), the
  strongest classical baseline on this design; the dark bar is the candidate the
  workflow selects. Timing is plotted as a magnitude, so lower is better in every
  panel.
  The candidates differ by a factor of $2.9$ in TNS but by under $1\%$ in area and
  $4.2\%$ in power, and all five are DRC-clean. Every candidate is at least as good
  as RTL-MP in area and power, and four of five are better in TNS and WNS, so the
  advantage does not depend on which candidate is selected.}
    \vskip -6pt
  \label{fig:bp_be_variance}
\end{figure}

Fig.~\ref{fig:bp_be_variance} illustrates this behavior for \texttt{bp\_be} at $1{:}1$, using RTL-MP as the classical reference.
The five candidates span $-9.42$ to $-27.27$~ns TNS, a $2.9\times$ range, while area varies by less than $1\%$ and power by $4.2\%$; all five are DRC-clean.
Individual candidates can underperform the baseline, as expected in exploratory search, but the selection mechanism identifies \texttt{asym\_5\_3} as the strongest candidate, improving TNS, WNS, power, and area by $60.0\%$, $53.6\%$, $4.1\%$, and $1.0\%$, respectively.
\textbf{Importantly, the benefit comes from structured exploration rather than a single favorable sample}: four of five candidates already outperform RTL-MP on both timing metrics, while all five match or improve its power and area. 
This pattern is consistent across the benchmark, with PlaceReasoner-Beta achieving the best timing among DRC-clean baselines for all eight square designs without increasing power or area.

\begin{table}[t]
\centering
\scriptsize
\setlength{\tabcolsep}{5pt}
\caption{Which post-route metrics macro placement actually moves, aggregated
over the 16 benchmark cells of Table~\ref{tab:variance_by_cell}.
``range\%'' is the candidate spread $(\max-\min)/|\mathrm{mean}|$.
Timing varies by tens of percent between candidates of the same design, whereas
power and area vary by at most a few percent.}
  \vskip -6pt
\begin{tabular}{@{}l rrr r@{}}
\toprule
Metric & mean CV\% & median CV\% & max CV\% & mean range\% \\
\midrule
TNS (ns) & 80.36 & 72.83 & 190.36 & 170.35 \\
WNS (ns) & 43.72 & 19.36 & 152.96 & 89.83 \\
Power (W) & 1.56 & 1.16 & 5.28 & 3.18 \\
Area ($\mu$m$^2$) & 0.27 & 0.19 & 0.98 & 0.56 \\
\bottomrule
\end{tabular}
  \vskip -6pt
\label{tab:variance_summary}
\end{table}

\begin{table*}[t]
\centering
\scriptsize
\setlength{\tabcolsep}{4pt}
\caption{Run-to-run PPA variance of PlaceReasoner on PlaceReasoner-Bench.
Statistics are taken over the $N=5$ placement candidates the planner generates
for each design, using only those that complete detailed routing; ``routed''
counts them and ``clean'' how many of those are DRC-clean.
$\mathrm{CV}\%=100\,\mathrm{std}/|\mathrm{mean}|$.
Timing spreads widely across candidates while power and area stay within a few
percent, which is why candidate selection is part of the workflow.}
  \vskip -6pt
\begin{tabular}{@{}l c cc rr rr rr rr@{}}
\toprule
\multirow{2}{*}{Design} & \multirow{2}{*}{AR} & \multicolumn{2}{c}{Candidates} & \multicolumn{2}{c}{WNS (ns)} & \multicolumn{2}{c}{TNS (ns)} & \multicolumn{2}{c}{Power (mW)} & \multicolumn{2}{c}{Area ($10^3\mu$m$^2$)} \\
\cmidrule(lr){3-4}\cmidrule(lr){5-6}\cmidrule(lr){7-8}\cmidrule(lr){9-10}\cmidrule(lr){11-12}
 & & routed & clean & mean$\pm$std & CV\% & mean$\pm$std & CV\% & mean$\pm$std & CV\% & mean$\pm$std & CV\% \\
\midrule
\texttt{ariane133} & $1{:}1$ & 2/5 & 1/2 & $-0.190\pm0.240$ & 126.5 & $-234.0\pm330.4$ & 141.2 & $268.0\pm14.1$ & 5.3 & $732.5\pm5.2$ & 0.71 \\
\texttt{ariane81} & $1{:}1$ & 5/5 & 5/5 & $-0.066\pm0.019$ & 29.5 & $-12.7\pm13.4$ & 105.5 & $195.2\pm1.5$ & 0.8 & $693.5\pm1.2$ & 0.17 \\
\texttt{bp\_be} & $1{:}1$ & 5/5 & 5/5 & $-0.226\pm0.077$ & 34.1 & $-19.3\pm6.5$ & 33.7 & $120.6\pm2.1$ & 1.7 & $239.2\pm0.8$ & 0.34 \\
\texttt{bp\_fe} & $1{:}1$ & 3/5 & 3/3 & $-0.257\pm0.393$ & 153.0 & $-15.9\pm27.2$ & 171.5 & $144.0\pm5.2$ & 3.6 & $215.1\pm2.1$ & 0.98 \\
\texttt{swerv\_wrapper} & $1{:}1$ & 5/5 & 1/5 & $-0.522\pm0.070$ & 13.4 & $-480.6\pm56.3$ & 11.7 & $244.4\pm4.8$ & 2.0 & $648.5\pm1.7$ & 0.26 \\
\texttt{vga\_lcd} & $1{:}1$ & 4/5 & 2/4 & $-0.800\pm0.045$ & 5.7 & $-96.0\pm63.8$ & 66.4 & $181.8\pm2.1$ & 1.1 & $649.6\pm0.9$ & 0.14 \\
\texttt{ethernet} & $1{:}1$ & 5/5 & 5/5 & $-0.314\pm0.015$ & 4.8 & $-1.5\pm0.1$ & 4.7 & $14.1\pm0.1$ & 0.4 & $110.5\pm0.1$ & 0.05 \\
\texttt{VeriGPU} & $1{:}1$ & 2/5 & 2/2 & $-0.260\pm0.014$ & 5.4 & $-12.6\pm0.5$ & 3.7 & $92.0\pm1.2$ & 1.3 & $274.3\pm0.3$ & 0.10 \\
\addlinespace[2pt]
\texttt{ariane133} & $2{:}1$ & 5/5 & 3/5 & $-0.058\pm0.050$ & 85.7 & $-11.6\pm20.7$ & 179.1 & $288.4\pm3.4$ & 1.2 & $731.2\pm1.5$ & 0.20 \\
\texttt{ariane81} & $2{:}1$ & 5/5 & 3/5 & $-0.172\pm0.214$ & 124.2 & $-172.2\pm327.9$ & 190.4 & $194.4\pm2.1$ & 1.1 & $694.3\pm1.8$ & 0.26 \\
\texttt{bp\_be} & $2{:}1$ & 3/5 & 1/3 & $-0.277\pm0.067$ & 24.1 & $-25.7\pm8.2$ & 31.8 & $107.7\pm1.5$ & 1.4 & $240.9\pm0.6$ & 0.24 \\
\texttt{bp\_fe} & $2{:}1$ & 4/5 & 2/4 & $-0.152\pm0.083$ & 54.2 & $-2.5\pm3.6$ & 144.5 & $130.8\pm1.3$ & 1.0 & $218.3\pm0.9$ & 0.41 \\
\texttt{swerv\_wrapper} & $2{:}1$ & 5/5 & 0/5 & $-0.586\pm0.086$ & 14.7 & $-534.0\pm50.4$ & 9.4 & $244.0\pm1.2$ & 0.5 & $648.2\pm0.8$ & 0.13 \\
\texttt{vga\_lcd} & $2{:}1$ & 5/5 & 2/5 & $-0.830\pm0.074$ & 8.9 & $-188.5\pm149.3$ & 79.2 & $182.8\pm1.9$ & 1.1 & $649.7\pm0.8$ & 0.13 \\
\texttt{ethernet} & $2{:}1$ & 5/5 & 5/5 & $-0.310\pm0.037$ & 12.1 & $-1.5\pm0.2$ & 11.6 & $13.9\pm0.0$ & 0.0 & $110.3\pm0.0$ & 0.04 \\
\texttt{VeriGPU} & $2{:}1$ & 5/5 & 2/5 & $-0.274\pm0.009$ & 3.3 & $-26.6\pm26.9$ & 101.4 & $129.2\pm3.3$ & 2.5 & $274.7\pm0.3$ & 0.10 \\
\bottomrule
\end{tabular}
  \vskip -6pt
\label{tab:variance_by_cell}
\end{table*}

% \subsubsection{Macro Orientation and Pin-Side Alignment.}

% \subsubsection{Optimization Trajectory of the Post-Route Optimizer.}

\subsection{Iterative Refinement: The Post-Route Optimization Trajectory}

\begin{figure}[t]
  \centering
  \begin{minipage}[b]{0.49\linewidth}
    \centering
    \includegraphics[width=\linewidth]{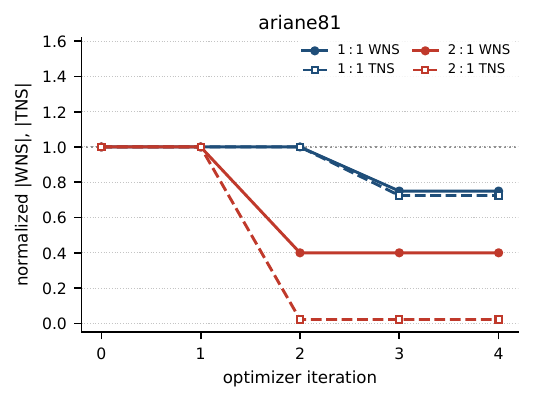}
  \end{minipage}\hfill
  \begin{minipage}[b]{0.49\linewidth}
    \centering
    \includegraphics[width=\linewidth]{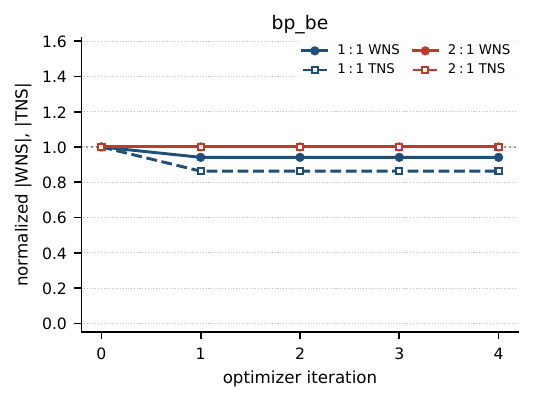}
  \end{minipage}\\[4pt]
  \begin{minipage}[b]{0.49\linewidth}
    \centering
    \includegraphics[width=\linewidth]{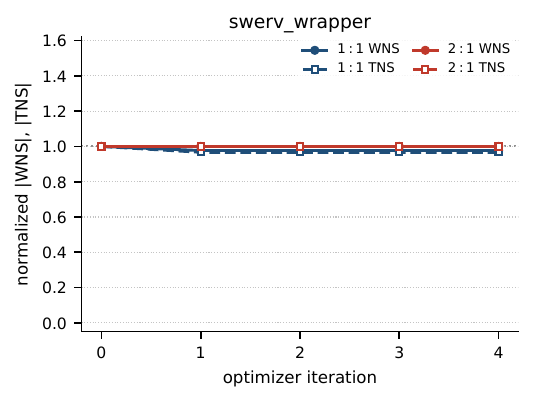}
  \end{minipage}\hfill
  \begin{minipage}[b]{0.49\linewidth}
    \centering
    \includegraphics[width=\linewidth]{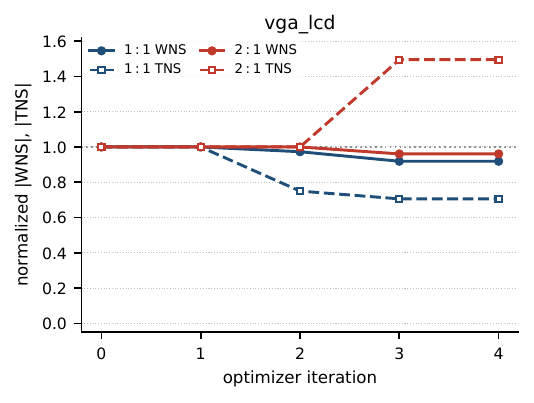}
  \end{minipage}
  \vskip -6pt
  \caption{Optimization trajectory of the post-route optimizer on four designs at
  both aspect ratios. Each curve follows the \emph{incumbent} placement, i.e.\ the
  candidate the workflow would ship after $k$ rounds, normalized by the base plan
  at iteration $0$; values below $1$ are improvements and both metrics are plotted
  as magnitudes. Color denotes the aspect ratio and line style the metric:
  solid with filled circles is WNS, dashed with open squares is TNS. A round that
  loses on timing or violates the DRC gate leaves the incumbent unchanged, which is
  why the curves are flat in places. The optimizer improves six of the eight runs,
  is unable to beat its base plan on \texttt{bp\_be} and \texttt{swerv\_wrapper} at
  $2{:}1$, and on \texttt{vga\_lcd} at $2{:}1$ improves WNS while degrading TNS.}
  \vskip -6pt
  \label{fig:opti_curve}
\end{figure}

The candidate search mentioned above concludes with the post-route optimizer, which receives routed feedback from the best base candidate and can perform up to four additional rounds of macro rearrangement. 
Fig.~\ref{fig:opti_curve} traces these rounds across four designs and both aspect ratios, showing the incumbent layout, the solution that would be selected after each iteration, normalized to the initial plan.
Only macro positions and orientations are modified; standard-cell placement, CTS, routing, and timing repair remain fixed, ensuring that improvements arise solely from macro-level refinement.
The optimizer improves the incumbent in \textbf{11 of 16 cases}, with gains often extending beyond the first round. 
For \texttt{vga\_lcd} at $1{:}1$, it widens inter-column routing channels from $4$ to $18,\mu$m over two rounds, reducing TNS by $29.4\%$ and WNS by $8.1\%$ while remaining DRC-clean. 
On \texttt{ariane81} at $2{:}1$, it discovers a four-corner arrangement transferred from a sibling design, improving WNS/TNS from $-0.05$/$-6.81$~ns to $-0.02$/$-0.15$~ns, a $97.8\%$ TNS reduction. 
Notably, this arrangement performed worse than the seed at $1{:}1$, illustrating that the agent can discover non-local alternatives that a purely local search would miss.

The trajectories also reveal clear limits.
On \texttt{bp\_be} and \texttt{swerv\_wrapper} at $2{:}1$, no refinement survives both timing and DRC checks, so the workflow retains the base plan.
On \texttt{vga\_lcd} at $2{:}1$, WNS improves by $3.9\%$, but TNS degrades by $49.5\%$, reflecting the current selection rule's WNS-first ranking rather than a failure to explore.
Two factors explain the diminishing returns. 
\textbf{First, geometric optimization is non-monotonic}: varying corridor position, width, channel width, or corner gaps often produces a local optimum near the initial arrangement, while larger moves can overshoot or violate legalization. 
\textbf{Second, macro placement has limited leverage when critical paths are macro-independent.} For example, the worst paths in both \texttt{swerv\_wrapper} runs are register-to-register paths between standard-cell blocks; macro placement can affect them only indirectly by reshaping the standard-cell region. 
Thus, post-route refinement provides substantial additional quality in many cases but cannot guarantee improvement. 
\textbf{A natural next step is critical-path-aware refinement that recognizes when macro-level changes have little remaining leverage and terminates exploration accordingly.}

% \subsubsection{Runtime and API Cost of PlaceReasoner-Beta.}

\subsection{Runtime and API Cost of PlaceReasoner-Beta}

The final two columns of Table~\ref{tab:pr_detail_ar1_4} quantify the computational cost of PlaceReasoner-Beta. 
Including $N=5$ candidate generation, checker verification, and iterative refinement, it requires $601.8$--$847.6$\,s per design (mean $733$\,s, $\sim$12 minutes), compared with $48.0$--$127.0$\,s for RTL-MP (mean $96$\,s) and $0.03$--$17$\,s for SA, DREAMPlace, and ChiPFormer. 
Token consumption ranges from $175.6$\,k on \texttt{bp\_fe} to $386.4$\,k on \texttt{vga\_lcd} (mean $290$\,k/design), increasing with macro count.
This overhead is modest relative to the physical-design loop.
\textbf{First, placement is a one-time cost}: a single downstream P\&R and timing-closure iteration can take hours, making $\sim$12 minutes a small fraction of the overall flow.
Moreover, PlaceReasoner-Beta does not require hundreds of optimization iterations to converge. 
\textbf{Second, runtime alone does not capture engineering cost.} 
SA completes in as little as $0.03$\,s but leaves $27{,}609$--$80{,}223$ DRC violations on six designs, while DREAMPlace's $5$--$11$\,s include three timeouts. 
A fast placement that requires substantial manual repair or fails to route provides little practical advantage.
Finally, API cost is a deployment choice rather than an inherent workflow cost: the same agent loop can run locally on an open-weight backbone with no per-token API charge.

\iffalse
The last two columns of Table~\ref{tab:pr_detail_ar1_4} report what the workflow
costs.
Including the generation of $N{=}5$ candidates, checker verification, and iterative
refinement, PlaceReasoner-Beta takes $601.8$--$847.6$\,s per design (mean $733$\,s,
about $12$ minutes), against $48.0$--$127.0$\,s for RTL-MP (mean $96$\,s) and
$0.03$--$17$\,s for SA, DREAMPlace, and ChiPFormer.
Token consumption, which is what the API is billed on, ranges from $175.6$\,k on
\texttt{bp\_fe} to $386.4$\,k on \texttt{vga\_lcd} (mean $290$\,k per design) and
grows with the number of macros the model must reason about.
Two considerations put this overhead in perspective.
First, it is a one-off, per-design cost paid \emph{before} the flow: a single
place-and-route and timing-closure iteration on these designs runs for hours, so
twelve minutes of macro placement is a small fraction of one iteration, and unlike
search-based methods PlaceReasoner-Beta does not require hundreds of such iterations to
converge.
Second, the cheap baselines are cheap only in isolation.
SA finishes in $0.03$\,s but leaves $27{,}609$--$80{,}223$ DRC violations on six of
the eight designs, and DREAMPlace's $5$--$11$\,s include three timeouts; a layout
that must be repaired by hand, or that never routes, costs far more engineering
time than the placement run it saved.
Finally, the API cost is a deployment choice rather than a property of the
workflow: the same agent loop runs on an open-weight backbone hosted locally, in
which case there is no per-token cost at all.

\fi

%; only a few minutes per design,

\subsection{Ablation Study}

Tables~\ref{tab:pr_detail_ar1_4} and~\ref{tab:pr_detail_ar2_4} report two controlled ablations: \textsc{CWI}, which removes layout images, and \textsc{CWS}, which disables the skill library. Both retain the full agent architecture and candidate-generation process. 
Their main impact is on \emph{iterative refinement}: without visual feedback, the model cannot inspect spatial errors, while without the skill library, it lacks explicit placement principles to guide corrections. 
Consequently, both variants tend to settle on early candidates; for example, \textsc{CWI} selects its first candidate on four of eight designs, compared with two for the full system.

On square tasks, removing either component generally degrades physical-design quality. Relative to PlaceReasoner-Beta, routed wirelength increases on six of eight designs (up to $17.7\%$ on \texttt{bp\_fe}), post-route area on seven, and power on five, with two ties. 
Timing remains competitive: the full system achieves better TNS on five of eight designs, while the ablations occasionally select a better-timed candidate, such as on \texttt{ariane81} and \texttt{swerv\_wrapper}. These cases highlight the role of best-of-five selection, where an ablation can occasionally benefit from a favorable sample. Nevertheless, the consistent wirelength, area, and power trends indicate that \textbf{visual grounding and encoded placement expertise provide broader and more stable benefits than timing optimization alone}.
The distinction becomes sharper on the elongated $2{:}1$ tasks, where limited routing space makes poor placement substantially harder to recover. 
\textsc{CWI} fails detailed routing on three designs, \texttt{ariane133}, \texttt{bp\_be}, and \texttt{bp\_fe}, while the full system routes all eight.
Without the layout image, the model cannot reliably recognize that its placement has narrowed or blocked critical routing channels. 
\textsc{CWS} completes routing on all designs but suffers severe timing degradation on the hardest cases, reaching $-2650.22$~ns TNS on \texttt{ariane133} versus $-0.08$~ns for the full system. 
This degradation reflects the absence of explicit peripheral-placement and cluster-locality principles needed to structure macros in constrained geometries.

\textbf{Overall, the ablations show that visual grounding and the skill library play complementary roles in the refinement loop: visual context enables the agent to perceive spatial defects, while encoded expertise provides actionable principles for correcting them.}
Because both are explicit framework components rather than learned parameters, they can be extended independently without retraining the underlying VLM.

\subsection{Discussion and Future Work}

PlaceReasoner-Beta achieves these results without task-specific training, using an agentic framework built around off-the-shelf VLMs, e.g., Claude-Opus-4.8 and the open-weight Qwen3-VL-8B/30B. 
This training-free property is also its main limitation and motivates the \emph{beta} designation: the three factors that most influence placement quality, i.e., model capability, exploration, and placement knowledge, remain largely external to the learning loop.
\noindent\textbf{Backbone.}
The open-weight models still occasionally generate overlapping or out-of-boundary macros, requiring Physical Checker-based repair and consuming refinement budget. 
A natural next step is to fine-tune a local model on a layout dataset, using supervised fine-tuning (SFT) for format and geometric validity followed by reinforcement learning (RL) with post-route rewards. This could improve placement generation while reducing reliance on proprietary APIs.
\noindent\textbf{Exploration.}
Candidates within a run remain highly correlated: even Claude-Opus-4.8 often produces structurally similar arrangements, limiting the effective search space of best-of-$N$ selection. 
Diversity-aware sampling or explicit search over candidate layouts could broaden exploration and increase the probability of discovering substantially better configurations.
\noindent\textbf{Placement knowledge.}
The skill pack explicitly encodes human placement principles, allowing the framework to apply established strategies but also inheriting their coverage and blind spots. 
Learning placement strategies directly from post-route feedback, e.g., through RL using the existing checker and physical-design rewards, could enable the system to discover strategies beyond those encoded by its designers.
\textbf{Closing these three gaps, i.e., a stronger locally trained backbone, more deliberate exploration, and placement knowledge learned from physical-design feedback, defines the path from the current beta framework toward a fully learning-driven agentic placer.}

\section{Conclusion}

We introduced \textbf{PlaceReasoner-Beta}, a multi-agent framework that reframes VLSI macro placement as a visual reasoning problem rather than black-box optimization.
A vision-language planner generates candidate layouts, geometric and physical checkers verify and refine them using placement principles and downstream feedback, and a post-route optimizer further improves promising solutions using final PPA.
We also introduced PlaceReasoner-Bench, a fully open end-to-end benchmark comprising 16 tasks with fixed floorplans and I/O assignments, ensuring that methods differ only in macro positions and orientations and are evaluated using routed PPA and DRC. 
Across the benchmark, PlaceReasoner-Beta achieves the best timing among DRC-clean methods on every square task and is the only approach to route all elongated tasks with at most two DRC violations. 
It also improves routed wirelength despite never explicitly optimizing it.
Because its reasoning, placement knowledge, and verification mechanisms are explicit framework components rather than learned parameters, PlaceReasoner-Beta is directly inspectable, extensible, and correctable without retraining. 
Together, PlaceReasoner-Beta and PlaceReasoner-Bench provide a foundation for developing and systematically evaluating reasoning-driven approaches to next-generation VLSI physical design.

\bibliographystyle{ACM-Reference-Format}
\bibliography{references}

@String{Computer = "{IEEE} Computer" }

@String{Springer = "Springer-Verlag" }

@INPROCEEDINGS{dreamplace,
  author={Liao, Peiyu and Liu, Siting and Chen, Zhitang and Lv, Wenlong and Lin, Yibo and Yu, Bei},
  booktitle={2022 Design, Automation \& Test in Europe Conference \& Exhibition (DATE)}, 
  title={DREAMPlace 4.0: Timing-driven Global Placement with Momentum-based Net Weighting}, 
  year={2022},
  volume={},
  number={},
  pages={939-944},
  doi={10.23919/DATE54114.2022.9774725}}

@inproceedings{rtl_mp,
  title={RTL-MP: toward practical, human-quality chip planning and macro placement},
  author={Kahng, Andrew B and Varadarajan, Ravi and Wang, Zhiang},
  booktitle={Proceedings of the 2022 International Symposium on Physical Design},
  pages={3--11},
  year={2022}
}

@article{maskplace,
  title={Maskplace: Fast chip placement via reinforced visual representation learning},
  author={Lai, Yao and Mu, Yao and Luo, Ping},
  journal={Advances in Neural Information Processing Systems},
  volume={35},
  pages={24019--24030},
  year={2022}
}

@inproceedings{chipformer,
  title={Chipformer: Transferable chip placement via offline decision transformer},
  author={Lai, Yao and Liu, Jinxin and Tang, Zhentao and Wang, Bin and Hao, Jianye and Luo, Ping},
  booktitle={International Conference on Machine Learning},
  pages={18346--18364},
  year={2023},
  organization={PMLR}
}

@book{vlsi_phy_design,
  title={VLSI physical design: from graph partitioning to timing closure},
  author={Kahng, Andrew B and Lienig, Jens and Markov, Igor L and Hu, Jin},
  volume={312},
  year={2011},
  publisher={Springer}
}

@misc{macro_guide,
  author       = {Swasti Pujari},
  title        = {Macro Placement Guidelines},
  year         = {2020},
  month        = aug,
  howpublished = {iVLSI Technologies},
  note         = {Accessed: July 21, 2026}
}

@article{macro_v1,
  title={NTUplace3: An analytical placer for large-scale mixed-size designs with preplaced blocks and density constraints},
  author={Chen, Tung-Chieh and Jiang, Zhe-Wei and Hsu, Tien-Chang and Chen, Hsin-Chen and Chang, Yao-Wen},
  journal={IEEE Transactions on Computer-Aided Design of Integrated Circuits and Systems},
  volume={27},
  number={7},
  pages={1228--1240},
  year={2008},
  publisher={IEEE}
}

@article{macro_v2,
  title={Replace: Advancing solution quality and routability validation in global placement},
  author={Cheng, Chung-Kuan and Kahng, Andrew B and Kang, Ilgweon and Wang, Lutong},
  journal={IEEE Transactions on Computer-Aided Design of Integrated Circuits and Systems},
  volume={38},
  number={9},
  pages={1717--1730},
  year={2018},
  publisher={IEEE}
}

@inproceedings{macro_v3,
  title={Reinforcement learning within tree search for fast macro placement},
  author={Geng, Zijie and Wang, Jie and Liu, Ziyan and Xu, Siyuan and Tang, Zhentao and Yuan, Mingxuan and Hao, Jianye and Zhang, Yongdong and Wu, Feng},
  booktitle={Forty-first International Conference on Machine Learning},
  year={2024}
}

@article{nature,
  title={A graph placement methodology for fast chip design},
  author={Mirhoseini, Azalia and Goldie, Anna and Yazgan, Mustafa and Jiang, Joe Wenjie and Songhori, Ebrahim and Wang, Shen and Lee, Young-Joon and Johnson, Eric and Pathak, Omkar and Nova, Azade and others},
  journal={Nature},
  volume={594},
  number={7862},
  pages={207--212},
  year={2021},
  publisher={Nature Publishing Group UK London}
}

@article{macro_v4,
  title={Macro placement by wire-mask-guided black-box optimization},
  author={Shi, Yunqi and Xue, Ke and Lei, Song and Qian, Chao},
  journal={Advances in Neural Information Processing Systems},
  volume={36},
  pages={6825--6843},
  year={2023}
}

@inproceedings{macro_v5,
  title={B*-trees: A new representation for non-slicing floorplans},
  author={Chang, Yun-Chih and Chang, Yao-Wen and Wu, Guang-Ming and Wu, Shu-Wei},
  booktitle={Proceedings of the 37th annual design automation conference},
  pages={458--463},
  year={2000}
}

@misc{CLIP,
      title={Learning Transferable Visual Models From Natural Language Supervision}, 
      author={Alec Radford and Jong Wook Kim and Chris Hallacy and Aditya Ramesh and Gabriel Goh and Sandhini Agarwal and Girish Sastry and Amanda Askell and Pamela Mishkin and Jack Clark and Gretchen Krueger and Ilya Sutskever},
      year={2021},
      eprint={2103.00020},
      archivePrefix={arXiv},
      primaryClass={cs.CV},
      url={https://arxiv.org/abs/2103.00020}, 
}

@article{qwen2-vl,
  title={Qwen2-vl: Enhancing vision-language model's perception of the world at any resolution},
  author={Wang, Peng and Bai, Shuai and Tan, Sinan and Wang, Shijie and Fan, Zhihao and Bai, Jinze and Chen, Keqin and Liu, Xuejing and Wang, Jialin and Ge, Wenbin and others},
  journal={arXiv preprint arXiv:2409.12191},
  year={2024}
}

@inproceedings{spatialvlm,
  title={Spatialvlm: Endowing vision-language models with spatial reasoning capabilities},
  author={Chen, Boyuan and Xu, Zhuo and Kirmani, Sean and Ichter, Brain and Sadigh, Dorsa and Guibas, Leonidas and Xia, Fei},
  booktitle={Proceedings of the IEEE/CVF Conference on Computer Vision and Pattern Recognition},
  pages={14455--14465},
  year={2024}
}

@article{internvl3,
  title={Internvl3. 5: Advancing open-source multimodal models in versatility, reasoning, and efficiency},
  author={Wang, Weiyun and Gao, Zhangwei and Gu, Lixin and Pu, Hengjun and Cui, Long and Wei, Xingguang and Liu, Zhaoyang and Jing, Linglin and Ye, Shenglong and Shao, Jie and others},
  journal={arXiv preprint arXiv:2508.18265},
  year={2025}
}

@book{physical_design,
  title={VLSI physical design automation: theory and practice},
  author={Sait, Sadiq M and Youssef, Habib},
  volume={6},
  year={1999},
  publisher={World Scientific Publishing Company}
}

@article{hier_rtl,
  title={Hier-RTLMP: A hierarchical automatic macro placer for large-scale complex IP blocks},
  author={Kahng, Andrew B and Varadarajan, Ravi and Wang, Zhiang},
  journal={IEEE Transactions on Computer-Aided Design of Integrated Circuits and Systems},
  volume={43},
  number={5},
  pages={1552--1565},
  year={2023},
  publisher={IEEE}
}

@inproceedings{challenge,
  title={Challenges in floorplanning and macro placement for modern SoCs},
  author={Tseng, I-Lun},
  booktitle={Proceedings of the 2024 International Symposium on Physical Design},
  pages={71--72},
  year={2024}
}

@inproceedings{openroad,
  title={Openroad: Toward a self-driving, open-source digital layout implementation tool chain},
  author={Ajayi, Tutu and Blaauw, David},
  booktitle={Proceedings of Government Microcircuit Applications and Critical Technology Conference},
  year={2019}
}

@misc{Nangate45,
    author       = {{NC State EDA}},
    title        = {{FreePDK45}},
    year         = {2011},
    howpublished = {\url{https://eda.ncsu.edu/freepdk/freepdk45/}},
    note         = {Version 1.4; accessed July 22, 2026}
  }

@inproceedings{sa,
  title={Assessment of reinforcement learning for macro placement},
  author={Cheng, Chung-Kuan and Kahng, Andrew B and Kundu, Sayak and Wang, Yucheng and Wang, Zhiang},
  booktitle={Proceedings of the 2023 International Symposium on Physical Design},
  pages={158--166},
  year={2023}
}

@article{layout-3d,
  title={Direct numerical layout generation for 3d indoor scene synthesis via spatial reasoning},
  author={Ran, Xingjian and Li, Yixuan and Xu, Linning and Yu, Mulin and Dai, Bo},
  journal={Advances in Neural Information Processing Systems},
  volume={38},
  pages={125055--125081},
  year={2026}
}

@article{chipbench,
  title={Benchmarking end-to-end performance of ai-based chip placement algorithms},
  author={Wang, Zhihai and Geng, Zijie and Tu, Zhaojie and Wang, Jie and Qian, Yuxi and Xu, Zhexuan and Liu, Ziyan and Xu, Siyuan and Tang, Zhentao and Kai, Shixiong and others},
  journal={Advances in Neural Information Processing Systems},
  volume={38},
  year={2026}
}

@article{pdagent,
  title={PDAGENT-BENCH: Characterizing, Grounding, and Architecting LLM/VLM Agents for VLSI Physical Design},
  author={Li, Qiufeng and Chen, Rongqian and Cheng, Quan and Wang, Chengxuan and Tang, Sizhe and Ho, Chia-Tung and Pan, David Z and Lan, Tian and Cao, Weidong},
  journal={arXiv preprint arXiv:2606.17253},
  year={2026}
}

@article{mage,
  title={MAGE: Human-Like Macro Placement via Agentic Multimodal Reasoning},
  author={Kahng, Andrew B and Kundu, Sayak and Pramanik, Bodhisatta},
  journal={arXiv preprint arXiv:2607.18536},
  year={2026}
}

@article{veoplace,
  title={See it to Place it: Evolving Macro Placements with Vision-Language Models},
  author={Uchendu, Ikechukwu and Goel, Swati and Hou, Karly and Songhori, Ebrahim and Lee, Kuang-Huei and Jiang, Joe Wenjie and Reddi, Vijay Janapa and Zhuang, Vincent},
  journal={arXiv preprint arXiv:2603.28733},
  year={2026}
}

@article{qwen3,
  title={Qwen3-vl technical report},
  author={Bai, Shuai and Cai, Yuxuan and Chen, Ruizhe and Chen, Keqin and Chen, Xionghui and Cheng, Zesen and Deng, Lianghao and Ding, Wei and Gao, Chang and Ge, Chunjiang and others},
  journal={arXiv preprint arXiv:2511.21631},
  year={2025}
}

@misc{opus48,
    title        = {Introducing Claude Opus 4.8},
    author       = {{Anthropic}},
    year         = {2026},
    month        = may,
    day          = {28},
    howpublished = {\url{https://www.anthropic.com/news/claude-opus-4-8}},
    note         = {Accessed: 2026-07-29}
  }

@inproceedings{yosys,
  title={Yosys-a free verilog synthesis suite},
  author={Wolf, Clifford and Glaser, Johann and Kepler, Johannes},
  booktitle={Proceedings of the 21st Austrian Workshop on Microelectronics (Austrochip)},
  volume={97},
  pages={1--6},
  year={2013}
}

@inproceedings{hierarchy_v1,
  title={Design-hierarchy aware mixed-size placement for routability optimization},
  author={Chuang, Yi-Lin and Nam, Gi-Joon and Alpert, Charles J and Chang, Yao-Wen and Roy, Jarrod and Viswanathan, Natarajan},
  booktitle={2010 IEEE/ACM International Conference on Computer-Aided Design (ICCAD)},
  pages={663--668},
  year={2010},
  organization={IEEE}
}

@inproceedings{hiercrchy_v2,
  title={Hierarchical global floorplacement using simulated annealing and network flow area migration},
  author={Choi, Wonjoon and Bazargan, Kia},
  booktitle={2003 Design, Automation and Test in Europe Conference and Exhibition},
  pages={1104--1105},
  year={2003},
  organization={IEEE}
}

@inproceedings{hierarchy_v3,
  title={Profile-guided microarchitectural floorplanning for deep submicron processor design},
  author={Ekpanyapong, Mongkol and Minz, Jacob R and Watewai, Thaisiri and Lee, Hsien-Hsin S and Lim, Sung Kyu},
  booktitle={Proceedings of the 41st annual Design Automation Conference},
  pages={634--639},
  year={2004}
}

@inproceedings{hierarchy_v4,
  title={Microarchitecture-aware floorplanning using a statistical design of experiments approach},
  author={Nookala, Vidyasagar and Chen, Ying and Lilja, David J and Sapatnekar, Sachin S},
  booktitle={Proceedings of the 42nd annual Design Automation Conference},
  pages={579--584},
  year={2005}
}

@inproceedings{autodmp,
  title={Autodmp: Automated dreamplace-based macro placement},
  author={Agnesina, Anthony and Rajvanshi, Puranjay and Yang, Tian and Pradipta, Geraldo and Jiao, Austin and Keller, Ben and Khailany, Brucek and Ren, Haoxing},
  booktitle={Proceedings of the 2023 International Symposium on Physical Design},
  pages={149--157},
  year={2023}
}

@article{eplace-ms,
  title={ePlace-MS: Electrostatics-based placement for mixed-size circuits},
  author={Lu, Jingwei and Zhuang, Hao and Chen, Pengwen and Chang, Hongliang and Chang, Chin-Chih and Wong, Yiu-Chung and Sha, Lu and Huang, Dennis and Luo, Yufeng and Teng, Chin-Chi and others},
  journal={IEEE Transactions on Computer-Aided Design of Integrated Circuits and Systems},
  volume={34},
  number={5},
  pages={685--698},
  year={2015},
  publisher={IEEE}
}

@article{ntuplace4h,
  title={NTUplace4h: A novel routability-driven placement algorithm for hierarchical mixed-size circuit designs},
  author={Hsu, Meng-Kai and Chen, Yi-Fang and Huang, Chau-Chin and Chou, Sheng and Lin, Tzu-Hen and Chen, Tung-Chieh and Chang, Yao-Wen},
  journal={IEEE Transactions on Computer-Aided Design of Integrated Circuits and Systems},
  volume={33},
  number={12},
  pages={1914--1927},
  year={2014},
  publisher={IEEE}
}

@article{effective_todaes,
  title={An effective floorplan-guided placement algorithm for large-scale mixed-size designs},
  author={Yan, Jackey Z and Viswanathan, Natarajan and Chu, Chris},
  journal={ACM Transactions on Design Automation of Electronic Systems (TODAES)},
  volume={19},
  number={3},
  pages={1--25},
  year={2014},
  publisher={ACM New York, NY, USA}
}

@inproceedings{learn_local,
  title={Learn to floorplan through acquisition of effective local search heuristics},
  author={He, Zhuolun and Ma, Yuzhe and Zhang, Lu and Liao, Peiyu and Wong, Ngai and Yu, Bei and Wong, Martin DF},
  booktitle={2020 IEEE 38th International Conference on Computer Design (ICCD)},
  pages={324--331},
  year={2020},
  organization={IEEE}
}

@inproceedings{cogagent,
  title={Cogagent: A visual language model for gui agents},
  author={Hong, Wenyi and Wang, Weihan and Lv, Qingsong and Xu, Jiazheng and Yu, Wenmeng and Ji, Junhui and Wang, Yan and Wang, Zihan and Dong, Yuxiao and Ding, Ming and others},
  booktitle={2024 IEEE/CVF Conference on Computer Vision and Pattern Recognition (CVPR)},
  pages={14281--14290},
  year={2024},
  organization={IEEE}
}

@article{gpt_web,
  title={Gpt-4v (ision) is a generalist web agent, if grounded},
  author={Zheng, Boyuan and Gou, Boyu and Kil, Jihyung and Sun, Huan and Su, Yu},
  journal={arXiv preprint arXiv:2401.01614},
  year={2024}
}

@article{ui-tars,
  title={Ui-tars: Pioneering automated gui interaction with native agents},
  author={Qin, Yujia and Ye, Yining and Fang, Junjie and Wang, Haoming and Liang, Shihao and Tian, Shizuo and Zhang, Junda and Li, Jiahao and Li, Yunxin and Huang, Shijue and others},
  journal={arXiv preprint arXiv:2501.12326},
  year={2025}
}

@article{large_survey,
  title={Large multimodal agents: A survey},
  author={Xie, Junlin and Chen, Zhihong and Zhang, Ruifei and Li, Guanbin},
  journal={Visual Intelligence},
  volume={3},
  number={1},
  pages={24},
  year={2025},
  publisher={Springer}
}

\section{Appendix}

\noindent\textbf{Full results on PlaceReasoner-Bench.}
Tables~\ref{tab:pr_detail_ar1_full} and~\ref{tab:pr_detail_ar2} give the complete
post-route results for all eight designs at the two aspect ratios, of which four
representative designs are reproduced in the main paper
(Tables~\ref{tab:pr_detail_ar1_4} and~\ref{tab:pr_detail_ar2_4}).
Every per-design count and average reported in the experiments is computed over
these full tables.

\begin{table*}[p]
\centering
\scriptsize
\setlength{\tabcolsep}{3pt}
\caption{Detailed post-route results for the square ($1{:}1$) tasks. ``--''
denotes an unavailable metric; failure states indicate that detailed routing
did not complete. Winner \# reports the one-based winner position $k/N$; use $N=5$; Runtime reports macro-placement time for
the classical baselines, summed model-driver time for the agent variants. Token Usage is the
total number of model tokens. }
  \vskip -6pt
\begin{tabular}{llcrrrrrrrcc}
\toprule
Design & Method & Winner \# & Route & WNS (ns) & TNS (ns) & DRC & WL ($\mu$m) & Power (W) & Area ($\mu$m$^2$) & Runtime (s)& Token Usage (k) \\
\midrule
ariane133 & RTL-MP & 2/5 & Yes & -0.10 & -4.60 & 0 & 5233982 & 0.265 & 729834 & 127.0 & -- \\
ariane133 & DREAMPlace & 3/5 & Fail: timeout & -- & -- & -- & -- & -- & -- & 11.00 & -- \\
ariane133 & ChiPFormer & 1/5 & Yes & -0.04 & -4.61 & 0 & 6211024 & 0.304 & 737387 & 16.96 & -- \\
ariane133 & SA & -- & Yes & -76.52 & -579878 & 80223 & 7135797 & 0.286 & 718246 & 0.240 & -- \\
ariane133 & CWI & 1/5 & Yes & -0.02 & -0.52 & 0 & 6105080 & 0.266 & 733327 & 814.4 & 329.4 \\
ariane133 & CWS & 5/5 & Yes & -0.02 & -0.56 & 0 & 5997840 & 0.265 & 732096 & 820.3 & 314.1 \\
ariane133 & PlaceReasoner-Beta & 3/5 & Yes & -0.02 & -0.35 & 0 & 5622394 & 0.258 & 728851 & 828.7 & 348.6 \\
ariane133 & Qwen3-VL-8B & -- & Fail: DRT-0073 & -- & -- & -- & -- & -- & -- & 3123.4 & 689.2 \\
ariane133 & Qwen3-VL-30B & 2/5 & Yes & -0.02 & -0.13 & 18166 & 6123046 & 0.283 & 728261 & 255.0 & 572.7 \\
ariane81 & RTL-MP & 2/5 & Yes & -0.14 & -8.43 & 0 & 5577716 & 0.184 & 694809 & 91.8 & -- \\
ariane81 & DREAMPlace & 1/5 & Fail: timeout & -- & -- & -- & -- & -- & -- & 9.53 & -- \\
ariane81 & ChiPFormer & 5/5 & Yes & -1.02 & -2103 & 0 & 6948313 & 0.177 & 700271 & 15.12 & -- \\
ariane81 & SA & -- & Fail: placed-only & -- & -- & -- & -- & -- & -- & 0.178 & -- \\
ariane81 & CWI & 1/5 & Yes & 0.00 & 0.00 & 0 & 4987790 & 0.171 & 693096 & 829.7 & 349.7 \\
ariane81 & CWS & 1/5 & Yes & 0.00 & 0.00 & 0 & 4987790 & 0.171 & 693096 & 836.1 & 337.2 \\
ariane81 & PlaceReasoner-Beta & 5/5 & Yes & -0.04 & -1.20 & 0 & 5203922 & 0.171 & 692142 & 847.6 & 362.6 \\
ariane81 & Qwen3-VL-8B & 2/5 & Yes & -0.09 & -6.24 & 0 & 6631628 & 0.198 & 693184 & 2449.0 & 759.3 \\
ariane81 & Qwen3-VL-30B & 1/5 & Yes & -0.09 & -13.95 & 0 & 6511947 & 0.203 & 694924 & 235.3 & 602.4 \\
bp\_be & RTL-MP & 4/5 & Yes & -0.28 & -23.54 & 0 & 2354232 & 0.123 & 240519 & 85.9 & -- \\
bp\_be & DREAMPlace & 4/5 & Yes & -0.47 & -29.81 & 63393 & 3814816 & -- & 243381 & 5.94 & -- \\
bp\_be & ChiPFormer & 1/5 & Yes & -0.58 & -54 & 0 & 2929620 & 0.131 & 243411 & 5.75 & -- \\
bp\_be & SA & -- & Yes & -0.24 & -21.26 & 37695 & 2411088 & 0.123 & 239997 & 0.032 & -- \\
bp\_be & CWI & 3/5 & Yes & -0.26 & -22.75 & 0 & 2598380 & 0.125 & 241666 & 693.7 & 198.5 \\
bp\_be & CWS & 3/5 & Yes & -0.26 & -22.75 & 0 & 2598379 & 0.125 & 241666 & 704.5 & 183.7 \\
bp\_be & PlaceReasoner-Beta & 1/5 & Yes & -0.13 & -9.42 & 0 & 2287187 & 0.118 & 238089 & 722.3 & 208.4 \\
bp\_be & Qwen3-VL-8B & 3/5 & Yes & -0.29 & -28.59 & 0 & 2435295 & 0.107 & 238967 & 349.7 & 308.9 \\
bp\_be & Qwen3-VL-30B & 1/5 & Yes & -0.16 & -13.28 & 0 & 2368477 & 0.102 & 238576 & 63.1 & 253.5 \\
bp\_fe & RTL-MP & 4/5 & Yes & -0.04 & -0.16 & 0 & 1607291 & 0.143 & 215331 & 97.9 & -- \\
bp\_fe & DREAMPlace & 5/5 & Yes & -0.37 & -41.92 & 65340 & 2743022 & -- & 223308 & 5.14 & -- \\
bp\_fe & ChiPFormer & 1/5 & Yes & -0.46 & -35 & 0 & 1795024 & 0.158 & 223786 & 5.74 & -- \\
bp\_fe & SA & -- & Yes & -0.19 & -1.06 & 0 & 1671812 & 0.145 & 217188 & 0.032 & -- \\
bp\_fe & CWI & 1/5 & Yes & -0.08 & -0.11 & 0 & 1828997 & 0.150 & 218063 & 577.9 & 154.7 \\
bp\_fe & CWS & 2/5 & Yes & -0.08 & -0.11 & 0 & 1828997 & 0.150 & 218063 & 586.3 & 158.9 \\
bp\_fe & PlaceReasoner-Beta & 3/5 & Yes & -0.03 & -0.07 & 0 & 1553678 & 0.141 & 213874 & 601.8 & 175.6 \\
bp\_fe & Qwen3-VL-8B & 1/5 & Yes & -0.13 & -4.75 & 0 & 1550338 & 0.126 & 215049 & 412.7 & 276.9 \\
bp\_fe & Qwen3-VL-30B & 4/5 & Yes & -0.19 & -7.20 & 0 & 1710046 & 0.139 & 221628 & 187.9 & 373.5 \\
ethernet & RTL-MP & 4/5 & Yes & -0.31 & -1.46 & 14 & 730251 & 0.0142 & 110629 & 102.5 & -- \\
ethernet & DREAMPlace & 1/5 & Yes & -0.30 & -1.38 & 30699 & 1496848 & -- & 112452 & 7.21 & -- \\
ethernet & ChiPFormer & 2/5 & Yes & -0.30 & -1.50 & 0 & 719208 & 0.014 & 110765 & 10.60 & -- \\
ethernet & SA & -- & Yes & -0.30 & -1.35 & 27609 & 761707 & 0.0142 & 110754 & 0.096 & -- \\
ethernet & CWI & 4/5 & Yes & -0.30 & -1.41 & 0 & 759078 & 0.0142 & 110764 & 618.4 & 264.1 \\
ethernet & CWS & 4/5 & Yes & -0.30 & -1.41 & 0 & 759078 & 0.0142 & 110764 & 635.6 & 266.2 \\
ethernet & PlaceReasoner-Beta & 1/5 & Yes & -0.30 & -1.44 & 0 & 760262 & 0.0142 & 110597 & 643.7 & 295.9 \\
ethernet & Qwen3-VL-8B & 4/5 & Yes & -0.30 & -1.45 & 0 & 735008 & 0.0171 & 110435 & 2252.7 & 780.8 \\
ethernet & Qwen3-VL-30B & 2/5 & Yes & -0.28 & -1.31 & 0 & 784869 & 0.0179 & 110483 & 121.7 & 398.7 \\
swerv\_wrapper & RTL-MP & 4/5 & Yes & -0.52 & -424.43 & 0 & 3757454 & 0.239 & 648507 & 122.2 & -- \\
swerv\_wrapper & DREAMPlace & 3/5 & Fail: timeout & -- & -- & -- & -- & -- & -- & 7.20 & -- \\
swerv\_wrapper & ChiPFormer & 4/5 & Yes & -1.55 & -1384 & 0 & 4593545 & 0.246 & 651840 & 8.89 & -- \\
swerv\_wrapper & SA & -- & Yes & -0.51 & -421.84 & 76026 & 4238062 & 0.240 & 649332 & 0.060 & -- \\
swerv\_wrapper & CWI & 3/5 & Yes & -0.40 & -330.99 & 0 & 3777640 & 0.240 & 648588 & 794.2 & 314.3 \\
swerv\_wrapper & CWS & 3/5 & Yes & -0.40 & -330.99 & 0 & 3777845 & 0.240 & 648588 & 805.6 & 321.2 \\
swerv\_wrapper & PlaceReasoner-Beta & 5/5 & Yes & -0.42 & -409.13 & 0 & 3447721 & 0.243 & 647807 & 811.9 & 335.7 \\
swerv\_wrapper & Qwen3-VL-8B & 1/5 & Yes & -0.60 & -617.00 & 0 & 4067535 & 0.249 & 646917 & 2060.6 & 547.9 \\
swerv\_wrapper & Qwen3-VL-30B & 1/5 & Yes & -0.56 & -558.07 & 0 & 4103206 & 0.245 & 646256 & 92.9 & 309.4 \\
vga\_lcd & RTL-MP & 5/5 & Yes & -0.78 & -498.02 & 45 & 2471807 & 0.182 & 650863 & 48.0 & -- \\
vga\_lcd & DREAMPlace & 3/5 & Yes & -0.85 & -146.70 & 62598 & 3804114 & -- & 650838 & 8.56 & -- \\
vga\_lcd & ChiPFormer & 2/5 & Yes & -0.82 & -1180 & 0 & 2880834 & 0.198 & 655252 & 11.15 & -- \\
vga\_lcd & SA & -- & Yes & -0.81 & -36.16 & 67293 & 2391472 & 0.184 & 649634 & 0.104 & -- \\
vga\_lcd & CWI & 3/5 & Yes & -0.66 & -48.84 & 0 & 2604281 & 0.184 & 650468 & 762.1 & 371.7 \\
vga\_lcd & CWS & 4/5 & Yes & -0.66 & -48.84 & 0 & 2604281 & 0.184 & 650468 & 774.9 & 364.1 \\
vga\_lcd & PlaceReasoner-Beta & 2/5 & Yes & -0.74 & -37.73 & 0 & 2342025 & 0.18 & 651477 & 785.1 & 386.4 \\
vga\_lcd & Qwen3-VL-8B & 4/5 & Yes & -0.89 & -41.92 & 0 & 2449050 & 0.190 & 648133 & 2084.3 & 730.6 \\
vga\_lcd & Qwen3-VL-30B & 2/5 & Yes & -0.71 & -27.86 & 0 & 2455068 & 0.192 & 647852 & 170.9 & 466.3 \\
VeriGPU & RTL-MP & 4/5 & Yes & -0.30 & -12.76 & 0 & 1753666 & 0.0931 & 274654 & 96.1 & -- \\
VeriGPU & DREAMPlace & 3/5 & Yes & -0.26 & -13.21 & 51828 & 2322311 & -- & 274662 & 6.44 & -- \\
VeriGPU & ChiPFormer & 2/5 & Yes & -0.26 & -13 & 0 & 1807519 & 0.095 & 275147 & 6.30 & -- \\
VeriGPU & SA & -- & Yes & -0.32 & -13.55 & 61089 & 1751564 & 0.0943 & 275045 & 0.029 & -- \\
VeriGPU & CWI & 1/5 & Yes & -0.26 & -12.56 & 0 & 1721060 & 0.0933 & 274566 & 604.5 & 198.3 \\
VeriGPU & CWS & 1/5 & Yes & -0.26 & -12.56 & 0 & 1721110 & 0.0933 & 274566 & 614.8 & 192.4 \\
VeriGPU & PlaceReasoner-Beta & 4/5 & Yes & -0.25 & -12.31 & 0 & 1697586 & 0.0911 & 274109 & 625.7 & 205.6 \\
VeriGPU & Qwen3-VL-8B & 1/5 & Yes & -0.25 & -17.43 & 0 & 1705359 & 0.128 & 274236 & 386.4 & 326.3 \\
VeriGPU & Qwen3-VL-30B & 3/5 & Yes & -0.28 & -15.68 & 0 & 1810348 & 0.131 & 274467 & 65.0 & 242.8 \\
\bottomrule
\end{tabular}
  \vskip -6pt
\label{tab:pr_detail_ar1_full}
\end{table*}

\begin{table*}[p]
\centering
\scriptsize
\setlength{\tabcolsep}{3pt}
\caption{Detailed post-route results for the elongated ($2{:}1$) tasks. Use the same settings with square (1 : 1) tasks.}
  \vskip -6pt
\label{tab:pr_detail_ar2}
\begin{tabular}{llcrrrrrrrcc}
\toprule
Design & Method & Winner \# & Route & WNS (ns) & TNS (ns) & DRC & WL ($\mu$m) & Power (W) & Area ($\mu$m$^2$) & Runtime (s) & Token Usage (k) \\
\midrule
ariane133 & RTL-MP & 1/5 & Yes & -0.12 & -25.29 & 0 & 6033112 & 0.276 & 731337 & -- & -- \\
ariane133 & DREAMPlace & -- & Fail: timeout & -- & -- & -- & -- & -- & -- & -- & -- \\
ariane133 & ChiPFormer & 2/5 & Yes & -0.39 & -620 & 0 & 7079270 & 0.278 & 738872 & -- & -- \\
ariane133 & SA & -- & Fail: GP1 & -- & -- & -- & -- & -- & -- & -- & -- \\
ariane133 & CWI & -- & Fail: GRT-0116 & -- & -- & -- & -- & -- & -- & 820.5 & 344.9 \\
ariane133 & CWS & 1/5 & Yes & -1.99 & -2650.22 & 0 & -- & 0.283 & 746548 & 827.6 & 348.7 \\
ariane133 & PlaceReasoner-Beta & 4/5 & Yes & -0.02 & -0.08 & 0 & 5690597 & 0.285 & 730228 & 834.9 & 357.1 \\
ariane133 & Qwen3-VL-8B & -- & Fail: GRT-0116 & -- & -- & -- & -- & -- & -- & 2228.3 & 479.9 \\
ariane133 & Qwen3-VL-30B & -- & Fail: DRT-0073 & -- & -- & -- & -- & -- & -- & 208.9 & 600.6 \\
ariane81 & RTL-MP & 2/5 & Yes & -0.10 & -35.10 & 0 & 6180469 & 0.179 & 696239 & -- & -- \\
ariane81 & DREAMPlace & -- & Fail: timeout & -- & -- & -- & -- & -- & -- & -- & -- \\
ariane81 & ChiPFormer & 4/5 & Yes & -1.82 & -5617 & 0 & 9503525 & 0.181 & 727526 & -- & -- \\
ariane81 & SA & -- & Fail: capped & -- & -- & -- & -- & -- & -- & -- & -- \\
ariane81 & CWI & 2/5 & Yes & -0.03 & -4.19 & 0 & 5217260 & 0.169 & 692640 & 848.6 & 387.4 \\
ariane81 & CWS & 2/5 & Yes & -0.03 & -4.19 & 0 & 5217585 & 0.169 & 692640 & 854.7 & 397.3 \\
ariane81 & PlaceReasoner-Beta & 1/5 & Yes & -0.02 & -0.15 & 0 & 5798709 & 0.194 & 698754 & 861.2 & 401.7 \\
ariane81 & Qwen3-VL-8B & 1/5 & Yes & -0.40 & -310.10 & 28 & 7017977 & 0.202 & 694691 & 778.8 & 330.1 \\
ariane81 & Qwen3-VL-30B & 2/5 & Yes & -0.09 & -43.12 & 1 & 5923522 & 0.193 & 692020 & 524.7 & 579.2 \\
bp\_be & RTL-MP & 1/5 & Yes & -0.37 & -38.96 & 0 & 2652279 & 0.123 & 240762 & -- & -- \\
bp\_be & DREAMPlace & -- & Fail: timeout & -- & -- & -- & -- & -- & -- & -- & -- \\
bp\_be & ChiPFormer & -- & Fail: segfault & -- & -- & -- & -- & -- & -- & -- & -- \\
bp\_be & SA & -- & Yes & -0.36 & -36.62 & 34190 & 2749978 & 0.124 & 241710 & -- & -- \\
bp\_be & CWI & -- & Fail: GRT-0116 & -- & -- & -- & -- & -- & -- & 714.9 & 221.5 \\
bp\_be & CWS & 3/5 & Yes & -0.08 & -1.72 & 0 & 2854209 & 0.124 & 240975 & 716.3 & 226.7 \\
bp\_be & PlaceReasoner-Beta & 4/5 & Yes & -0.22 & -18.06 & 1 & 2873409 & 0.109 & 240882 & 728.9 & 241.6 \\
bp\_be & Qwen3-VL-8B & 1/5 & Yes & -0.39 & -31.03 & 0 & 2647452 & 0.107 & 239686 & 216.1 & 237.2 \\
bp\_be & Qwen3-VL-30B & 1/5 & Fail: GRT-0116 & -- & -- & -- & -- & -- & -- & 256.2 & 480.3 \\
bp\_fe & RTL-MP & 1/5 & Yes & -0.18 & -6.12 & 0 & 1609885 & 0.143 & 215198 & -- & -- \\
bp\_fe & DREAMPlace & -- & Fail: timeout & -- & -- & -- & -- & -- & -- & -- & -- \\
bp\_fe & ChiPFormer & 1/5 & Yes & -0.05 & -0.21 & 5 & 2286401 & 0.149 & 216837 & -- & -- \\
bp\_fe & SA & -- & Yes & -0.17 & -6.55 & 20479 & 1662195 & -- & 218725 & -- & -- \\
bp\_fe & CWI & -- & Fail: GRT-0116 & -- & -- & -- & -- & -- & -- & 603.8 & 164.8 \\
bp\_fe & CWS & 1/5 & Yes & -0.08 & -0.93 & 0 & 2227285 & 0.152 & 217744 & 614.1 & 166.2 \\
bp\_fe & PlaceReasoner-Beta & 2/5 & Yes & -0.11 & -0.23 & 0 & 1638383 & 0.131 & 217638 & 627.4 & 186.7 \\
bp\_fe & Qwen3-VL-8B & 2/5 & Yes & -0.09 & -2.43 & 0 & 1719403 & 0.135 & 219808 & 202.9 & 204.9 \\
bp\_fe & Qwen3-VL-30B & 3/5 & Yes & -0.11 & -1.19 & 0 & 1637963 & 0.126 & 215246 & 62.8 & 272.8 \\
ethernet & RTL-MP & 3/5 & Yes & -0.33 & -1.62 & 14 & 812042 & 0.014 & 110330 & -- & -- \\
ethernet & DREAMPlace & 1/5 & Yes & -0.30 & -1.38 & 34020 & 1383506 & -- & 112103 & -- & -- \\
ethernet & ChiPFormer & 1/5 & Yes & -0.33 & -1.60 & 0 & 797665 & 0.014 & 110403 & -- & -- \\
ethernet & SA & -- & Yes & -0.27 & -1.31 & 10394 & 867444 & -- & 110537 & -- & -- \\
ethernet & CWI & 4/5 & Yes & -0.29 & -1.35 & 0 & 849848 & 0.014 & 110644 & 665.8 & 314.2 \\
ethernet & CWS & 4/5 & Yes & -0.29 & -1.35 & 0 & 849848 & 0.014 & 110644 & 672.6 & 311.8 \\
ethernet & PlaceReasoner-Beta & 2/5 & Yes & -0.28 & -1.34 & 0 & 825144 & 0.0139 & 110333 & 684.2 & 327.5 \\
ethernet & Qwen3-VL-8B & 1/5 & Yes & -0.29 & -1.43 & 0 & 761143 & 0.017 & 110252 & 554.9 & 399.0 \\
ethernet & Qwen3-VL-30B & 1/5 & Yes & -0.26 & -1.27 & 0 & 796341 & 0.0172 & 110237 & 188.8 & 498.3 \\
swerv\_wrapper & RTL-MP & 1/5 & Yes & -0.64 & -528.28 & 1421 & 3779122 & 0.235 & 647771 & -- & -- \\
swerv\_wrapper & DREAMPlace & -- & Fail: timeout & -- & -- & -- & -- & -- & -- & -- & -- \\
swerv\_wrapper & ChiPFormer & 1/5 & Yes & -1.40 & -1616 & 1456 & 6958764 & 0.260 & 686355 & -- & -- \\
swerv\_wrapper & SA & -- & Fail: placed-only & -- & -- & -- & -- & -- & -- & -- & -- \\
swerv\_wrapper & CWI & 5/5 & Yes & -0.67 & -507.90 & 0 & 4176586 & 0.238 & 648265 & 688.1 & 407.2 \\
swerv\_wrapper & CWS & 4/5 & Yes & -0.67 & -507.90 & 0 & 4176586 & 0.238 & 648265 & 695.4 & 409.4 \\
swerv\_wrapper & PlaceReasoner-Beta & 1/5 & Yes & -0.58 & -476.28 & 2 & 3925621 & 0.242 & 648145 & 716.8 & 415.6 \\
swerv\_wrapper & Qwen3-VL-8B & 4/5 & Yes & -0.67 & -755.17 & 0 & 4471772 & 0.245 & 646254 & 335.9 & 214.1 \\
swerv\_wrapper & Qwen3-VL-30B & 2/5 & Yes & -0.82 & -927.46 & 63 & 4918313 & 0.248 & 647625 & 102.9 & 397.6 \\
vga\_lcd & RTL-MP & 1/5 & Yes & -0.72 & -29.70 & 171 & 2354087 & 0.180 & 647137 & -- & -- \\
vga\_lcd & DREAMPlace & 2/5 & Yes & -0.77 & -40.80 & 52227 & 3089394 & -- & 649017 & -- & -- \\
vga\_lcd & ChiPFormer & 1/5 & Yes & -0.88 & -2317 & 33 & 3688031 & 0.198 & 654877 & -- & -- \\
vga\_lcd & SA & -- & Fail: GP1 & -- & -- & -- & -- & -- & -- & -- & -- \\
vga\_lcd & CWI & 3/5 & Yes & -0.72 & -27.99 & 0 & 2545176 & 0.183 & 649913 & 725.9 & 314.9 \\
vga\_lcd & CWS & 2/5 & Yes & -0.72 & -27.99 & 0 & 2545176 & 0.183 & 649913 & 736.9 & 320.4 \\
vga\_lcd & PlaceReasoner-Beta & 3/5 & Yes & -0.76 & -33.13 & 1 & 2547982 & 0.183 & 649875 & 748.6 & 348.3 \\
vga\_lcd & Qwen3-VL-8B & 1/5 & Yes & -0.76 & -1154.16 & 25 & 2605124 & 0.187 & 648128 & 816.1 & 424.9 \\
vga\_lcd & Qwen3-VL-30B & 3/5 & Yes & -0.91 & -226.91 & 35 & 2750093 & 0.193 & 649266 & 125.7 & 468.3 \\
VeriGPU & RTL-MP & 1/5 & Yes & -0.30 & -16.98 & 22 & 2009432 & 0.091 & 274307 & -- & -- \\
VeriGPU & DREAMPlace & 2/5 & Yes & -0.49 & -19.39 & 62832 & 2727467 & -- & 274799 & -- & -- \\
VeriGPU & ChiPFormer & 1/5 & Yes & -0.28 & -17 & 0 & 2108273 & 0.093 & 275311 & -- & -- \\
VeriGPU & SA & -- & Yes & -0.28 & -17.51 & 45375 & 1996265 & -- & 275007 & -- & -- \\
VeriGPU & CWI & 5/5 & Yes & -0.29 & -14.13 & 0 & 1977510 & 0.0919 & 274578 & 313.5 & 211.7 \\
VeriGPU & CWS & 5/5 & Yes & -0.29 & -14.13 & 0 & 1977510 & 0.0919 & 274578 & 339.4 & 208.9 \\
VeriGPU & PlaceReasoner-Beta & 1/5 & Yes & -0.28 & -13.08 & 0 & 1896523 & 0.101 & 274461 & 367.1 & 225.6 \\
VeriGPU & Qwen3-VL-8B & 1/5 & Yes & -0.45 & -54.06 & 0 & 1906068 & 0.129 & 274392 & 227.8 & 327.8 \\
VeriGPU & Qwen3-VL-30B & 2/5 & Yes & -0.29 & -15.41 & 0 & 1931436 & 0.128 & 274339 & 199.8 & 389.8 \\
\bottomrule
\end{tabular}
  \vskip -6pt
\end{table*}

\end{document}